\documentclass[11pt]{article}

\usepackage{acl}
\usepackage{times}
\usepackage{latexsym}

\usepackage[T1]{fontenc}

\usepackage[utf8]{inputenc}

\usepackage{microtype}

\usepackage{inconsolata}

\usepackage{cases}
\usepackage{graphicx}
\usepackage{algpseudocode}

\usepackage{amsmath, amssymb, amsthm}
\usepackage[ruled,vlined,linesnumbered]{algorithm2e}
\newtheorem{theorem}{Theorem}

\usepackage{enumitem}
\usepackage{algpseudocode}
\usepackage{booktabs}
\usepackage{multirow}
\usepackage{booktabs}
\usepackage{multirow}
\usepackage{subcaption}
\usepackage{booktabs}  
\usepackage{makecell}  
\usepackage{graphicx}  
\usepackage{algpseudocode} 
\usepackage{booktabs}
\usepackage{multirow}
\usepackage[table]{xcolor} 
\usepackage{pgf}           

\newcommand{\heat}[1]{%
    \pgfmathparse{#1/0.26*100}
    \ifdim\pgfmathresult pt>100pt\def\temp{100}\else\edef\temp{\pgfmathresult}\fi
    \cellcolor{red!\temp}#1
}
\usepackage{booktabs,multirow,xcolor}
\title{ When Benchmarks Leak: Inference-Time Decontamination for LLMs}

\author{
  Chai Jianzhe \\
  Institute of Science Tokyo \\
  \texttt{chai.j.4b1c@m.isct.ac.jp}
  \And
  Yu Zhe \\
  RIKEN AIP \\
  \texttt{zhe.yu@riken.jp}
  \AND
  Jun Sakuma \\
  Institute of Science Tokyo \\
  RIKEN AIP \\
  \texttt{sakuma@c.titech.ac.jp}
}

\begin{document}
\maketitle
\begin{abstract}

Benchmark-based evaluation is the de facto standard for comparing large language models (LLMs). However, its reliability is increasingly threatened by test set contamination, where test samples or their close variants leak into training data and artificially inflate reported performance. To address this issue, prior work has explored two main lines of mitigation. One line attempts to identify and remove contaminated benchmark items before evaluation, but this inevitably alters the evaluation set itself and becomes unreliable when contamination is moderate or severe. The other line preserves the benchmark and instead suppresses contaminated behavior at evaluation time; however, such interventions often interfere with normal inference and lead to noticeable performance degradation on clean inputs. We propose DeconIEP, a decontamination framework that operates entirely during evaluation by applying small, bounded perturbations in the input embedding space. Guided by a relatively less-contaminated reference model, DeconIEP learns an instance-adaptive perturbation generator that steers the evaluated model away from memorization-driven shortcut pathways. Across multiple open-weight LLMs and benchmarks, extensive empirical results show that DeconIEP achieves strong decontamination effectiveness while incurring only minimal degradation in benign utility.

\end{abstract}
\section{Introduction}

\begin{figure}
    \centering
    \includegraphics[width=1\linewidth]{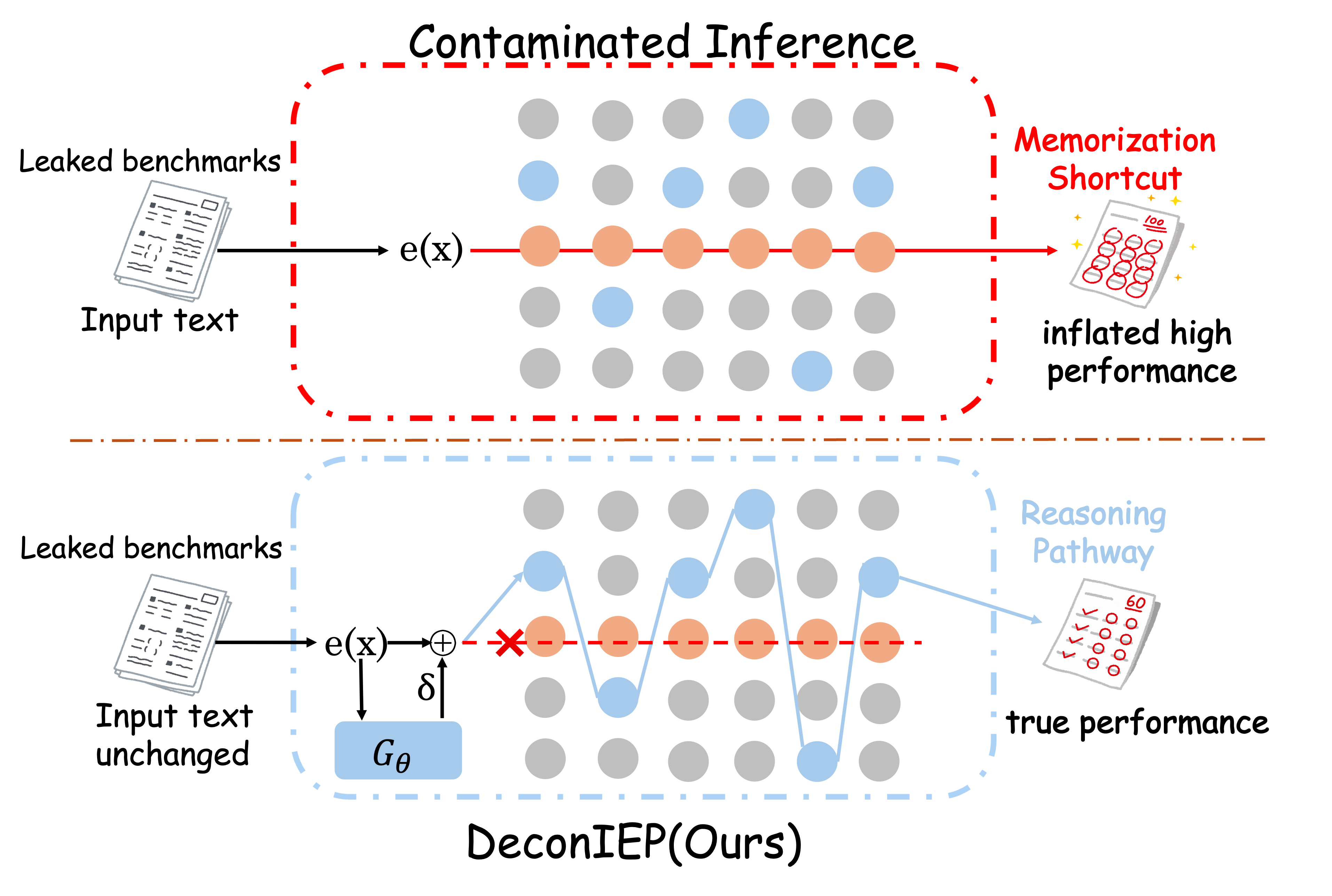}
\caption{Overview: DeconIEP keeps the prompt fixed and adds a bounded embedding perturbation $\delta=G_\theta(e(x))$ to suppress memorization shortcuts on leaked samples, steering inference toward reasoning and recovering clean performance.}

    \label{fig:intro}
\end{figure}
Large language models (LLMs) have reached a level of performance that enables them to solve a broad range of knowledge, reasoning, and code-related tasks \citep{liang2023holisticevaluationlanguagemodels}. As these models are increasingly developed, released, and deployed by different organizations, a systematic way to evaluate and compare their capabilities becomes necessary. In this setting, benchmark-based evaluation has naturally emerged as a widely adopted practice for assessing LLM performance. For example, prominent models such as GPT-5 \citep{openai_gpt5_2025} are often presented alongside their scores on established benchmarks like MMLU \citep{hendrycks2021measuringmassivemultitasklanguage}, where benchmark results serve as standardized evidence of general reasoning ability. 

Recently, the need for trustworthy evaluation is further amplified by the rapid growth of \emph{open-weight} LLMs. 
Unlike closed APIs, open-weight models can be freely downloaded, fine-tuned, merged, and re-distributed, enabling fast iteration by both industry and the open-source community.
As a result, benchmark scores are increasingly used not only for marketing, but also for model selection, deployment decisions, and reproducible scientific comparison across independently trained variants.
This makes reliable evaluation of open-weight models particularly important: their training pipelines and downstream fine-tuning are more decentralized, and thus more prone to unintended benchmark exposure and evaluation artifacts.

However, benchmark-centric evaluation is vulnerable to data contamination~\citep{deng2024unveilingspectrumdatacontamination,deng2024investigating,chen2025benchmarking}, where test items (or closely related variants) appear in training data thereby violating the train--test separation and inflating reported performance. This inflation can misrepresent real-world generalization. For example, it has been reported that over 15 LLMs exhibit inflated benchmark performance on six popular benchmarks, where the estimated contamination rate ranges from \(1\%\) to \(45\%\), and some of them are fine-tuned models derived from open-weight models \citep{li-etal-2024-open-source}.

To mitigate contamination, prior work has explored multiple strategies,
including detect-then-filter approaches based on membership inference attacks
(MIA)~\citep{salem2018mlleaksmodeldataindependent,hayes2018loganmembershipinferenceattacks}.
These methods apply an MIA detector to identify leaked benchmark items and
exclude them from evaluation. However, filtering contaminated samples fundamentally alters the benchmark,
making the resulting evaluation set no longer directly comparable to the
original benchmark used in prior work.
Moreover, MIA detectors are inherently imperfect: any non-zero false-negative
rate leaves residual leaked items unfiltered, which can continue to inflate
reported performance, especially under non-trivial contamination.
As shown in Appendix~\ref{app:mia_limitation}, we formally prove this
limitation and validate it empirically.

These limitations motivate inference-time decontamination~\citep{zhu2024inference,dong2024generalization,zhu2025establishingtrustworthyllmevaluation}, which mitigates contamination effects during model inference rather than by filtering benchmark items.
A common issue of existing methods is that the same intervention used for decontamination applied to contaminated inputs also affects clean ones: interventions strong enough to reduce contamination often alter model behavior on uncontaminated inputs, leading to non-trivial drops in clean accuracy.
For example, rewriting methods~\citep{zhu2024inference} may shift the effective input distribution, while internal-intervention approaches~\citep{zhu2025establishingtrustworthyllmevaluation,menta2025analyzing} can suppress internal signals that are useful for both contaminated and clean examples.
As a result, existing methods face an unfavorable trade-off between contamination mitigation and clean performance, motivating our goal of minimizing benign utility degradation.

To address this limitation, we propose a decontamination framework, termed DeconIEP (\textbf{Decon}tamination via \textbf{I}nference-time \textbf{E}mbedding \textbf{P}erturbation). 
We observe that data contamination mainly manifests as a change in how a model internally supports its predictions. On clean test samples, predictions are typically supported by broadly shared
reasoning features that generalize across many inputs. In contrast, as shown in the first row of Figure \ref{fig:intro}, on contaminated test samples, the same outputs can be produced by relying on a different set of contamination-specific components, such as
shortcut neurons or late-layer retrieval pathways, thereby reducing the need for genuine reasoning~\citep{zhu2025establishingtrustworthyllmevaluation, menta2025analyzing}.
This observation motivates our approach: as shown in the second row of Figure \ref{fig:intro}, by applying small, targeted perturbations in the embedding space, we weaken contamination-driven dependencies and steer the model toward a cleaner inference regime, while preserving the input representation.

Based on this insight, we train a generator to produce inference-time embedding perturbations, guided by a comparatively less contaminated reference model. Our setting targets models obtained via fine-tuning from a known pretrained checkpoint (e.g., Llama3), where benchmark contamination is typically introduced or amplified by fine-tuning the checkpoint. In this context, an earlier checkpoint or base model with the same architecture naturally provides a practical reference with reduced benchmark-specific exposure. Importantly, our empirical evaluation demonstrates that our approach does not require a perfectly clean reference model. It suffices that the reference exhibits relatively less contamination, allowing the generator to suppress contamination-driven shortcuts while preserving benign reasoning behavior, without modifying the parameters of the evaluated model.

\paragraph{Contributions}
Our contributions are two-fold: (1) We propose {DeconIEP}, a novel inference-time decontamination
    method that mitigates benchmark contamination by applying small,
    targeted perturbations to input embeddings, without modifying model
    parameters or the benchmark itself. (2) We conduct extensive experiments across multiple benchmarks and
    contamination settings, showing that DeconIEP effectively reduces
    contamination-induced performance inflation while incurring
    substantially smaller degradation on clean inputs compared to existing
    inference-time baselines.

\begin{table*}[htb]
\small
  \centering
  \resizebox{\textwidth}{!}{%
  \begin{tabular}{lccccc}
    \toprule
    \textbf{Method} 
    & \textbf{Setting} 
    & \textbf{Core Mechanism} 
    & \textbf{\makecell{Need Reference\\ Model}} 
    & \textbf{\makecell{Decontam. \\ Efficacy $\uparrow$}} 
    & \textbf{\makecell{Clean Eval \\ Drop $\downarrow$}} \\
    \midrule

    \textbf{TED} \citep{dong2024generalization} 
    & Black-box & Distribution calibration 
    & No
    & Medium & Medium \\

    \textbf{ITD} \citep{zhu2024inference} 
    & Black-box & Query rewriting 
    & No
    & Limited & \textbf{Low} \\

    \midrule

    \textbf{Shortcut Neuron} \citep{zhu2025establishingtrustworthyllmevaluation} 
    & White-box & Activation patching 
    & \textbf{Yes}
    & \textbf{High} & High \\

    \textbf{Short-circuiting} \citep{menta2025analyzing} 
    & White-box & Attention bypassing 
    & No
    & \textbf{High} & High \\


    \midrule
    \textbf{DeconIEP(Ours)} 
    & \textbf{White-box} & \textbf{Embedding perturbation} 
    & \textbf{Yes}
    & \textbf{High} & \textbf{Low} \\
    \bottomrule
  \end{tabular}%
  }
  \caption{\label{tab:comparison}
    Comparison of decontamination strategies.
    \textit{Decontam. Efficacy} indicates decontamination effectiveness on contaminated benchmarks.
    \textit{Clean Eval Drop} indicates performance degradation on non-contaminated tasks.
    \textit{Need Reference Model} indicates whether a reference model is required.
  }
\end{table*}

\section{Related Work}
\label{sec:related-work}

\paragraph{Benchmark Contamination and Detection}
Data contamination (test-set leakage) in widely used benchmarks (e.g., MMLU,
GSM8K) can inflate reported performance by inducing memorization rather than
genuine generalization \citep{chen2025benchmarking,deng2024investigating,xu2024benchmarking}.
When training data are accessible, overlap-based checks can reveal contamination
\citep{brown2020language,gao2021framework}, but for most open-source models such
verification is infeasible. An alternative line of work adopts detect-then-filter strategies based on
membership inference attacks (MIAs), but imperfect detectors leave residual
contamination and may introduce selection bias (see Appendix~\ref{app:mia_limitation}).

\paragraph{Inference-time Decontamination (Black-box)}
To avoid training-data access, API-only methods intervene at inference time:
TED~\citep{dong2024generalization} calibrates outputs via repeated sampling, and
ITD~\citep{zhu2024inference} rewrites prompts using auxiliary LLMs.
These approaches are easy to deploy but can be computationally expensive, and
prompt rewriting may alter semantics or task difficulty \citep{xu2024benchmarking}.

\paragraph{Inference-time Decontamination (White-box)}
With open-weight access, prior work manipulates internal representations,
Short-cut Neuron Analysis~\citep{zhu2025establishingtrustworthyllmevaluation} uses a reference model to identify such shortcut neurons and then edits their activations at test time.
As shown in Table~\ref{tab:comparison}, it can achieve high decontamination efficacy, but often leads to a large clean-evaluation drop when the edited neurons overlap with general-purpose computation.
Menta et al.~\citep{menta2025analyzing} propose {attention short-circuiting}, which replaces the attention mixing operation with an identity mapping so that value vectors are forwarded without cross-token aggregation.
They report that bypassing attention in deeper layers can substantially reduce the generation of memorized content.
While this mechanism is not proposed as a decontamination method, it can be repurposed to suppress contamination-driven memorization.
However, as summarized in Table~\ref{tab:comparison}, the same intervention also degrades general capabilities, leading to a large benign utility drop.


\section{Problem Setup}

\label{subsec:def}

\paragraph{Model}


Let $f: \mathcal{X} \rightarrow \mathcal{Y}$ be a Large Language Model (LLM), where $\mathcal{X}$ is the space of input sequences and $\mathcal{Y}$ is the space of output sequences.

\paragraph{Data Contamination}

We want to train a model $f$ and we have a test benchmark
$D_{\text{test}} \sim \mathcal{P}_{\text{test}}$ to evaluate its performance. We define the \emph{contaminated} model $f_{\mathrm{con}}$ as any model trained
on a dataset
\[
D_{\mathrm{train}}^{\mathrm{con}}
\;=\;
D_{\mathrm{train}}^{\mathrm{clean}}\,\cup D_{\mathrm{con}},
\]
where $D_{\mathrm{train}}^{\mathrm{clean}}$ is general training data and unrelated to $D_{\text{test}}$, and $D_{\mathrm{con}}$ potentially causes training data contamination, where we consider the following three levels of contamination as follows.

\begin{enumerate}
    \item \textbf{Exact Contamination}:  $D_{\mathrm{con}}  \subseteq D_{\text{test}}$, i.e., the
    training data contains exact copies of test examples.
    \item \textbf{Semantic-level Contamination}: 
    For each $x \in D_{\mathrm{test}}$, there exists $x' \in D_{\mathrm{con}}$ such that, $\mathrm{sem}(x)=\mathrm{sem}(x')$. Here,  $\mathrm{sem}(\cdot)$ refers to paraphrasing or semantically equivalent rewriting by LLM or human.
    \item \textbf{Domain-level Contamination}: $D_{\mathrm{con}} \sim \mathcal{P}_{\text{test}}$, i.e., the
    contamination set and the test set are drawn from (approximately) the same
    underlying distribution, such as $D_{\text{con}}$ and $D_{\text{test}}$ are two random splits of the same benchmark.
\end{enumerate}


For a fixed test benchmark $D_{\mathrm{test}}$, we call a model $f_{\mathrm{clean}}$ trained on $D_{\mathrm{train}}^{\mathrm{clean}}$ \emph{clean} if its training data does not contaminate $D_{\mathrm{test}}$ at any of three contamination level defined above.

Due to contamination, the observed performance of $f_{\text{con}}$ on $D_{\text{test}}$ is often inflated:
\begin{equation}
\text{Perf}(f_{\text{con}}, D_{\text{test}}) > \text{Perf}(f_{\text{clean}}, D_{\text{test}})
\end{equation}
Here, $\text{Perf}(\cdot)$  is a general performance evaluator, for example, classification accuracy for multiple-choice benchmarks (e.g., MMLU).

\paragraph{Objective}
We design an inference-time mitigation operator
\[
M:\ \mathcal{F}\rightarrow\mathcal{F}
\]
which maps a contaminated model $f_{\mathrm{con}}$ to a mitigated predictor $M(f_{\mathrm{con}})$ with a modified inference procedure.
Our objective is to make its test performance match that of an ideal clean model:
\begin{equation}
\mathrm{Perf}\bigl(M(f_{\mathrm{con}}), D_{\mathrm{test}}\bigr)\ \approx\ \mathrm{Perf}\bigl(f_{\mathrm{clean}}, D_{\mathrm{test}}\bigr),
\end{equation}

\section{Methodology}
\label{sec:method}

\subsection{Overview}
\label{subsec:method_overview}


We reduce contamination-induced inflation by correcting inference rather than filtering items.
Given $f_{\mathrm{con}}$ and input $x$, we keep the prompt unchanged and add a bounded embedding perturbation $\delta(x)$ with $\|\delta(x)\|_{\infty}\le \zeta$.
We train a generator $G_{\theta}$ via a KL objective (Theorem~\ref{thm:kl-bound}) to align $f_{\mathrm{con}}$ with a reference model $f_{\mathrm{ref}}$.
At test time, we apply $\delta(x)$ once per input and evaluate on the original benchmark.

\subsection{KL Divergence as a Surrogate Objective}
\label{subsec:kl-objective}

Our goal is to align the behavior of the mitigated contaminated model
$M(f_{\mathrm{con}}) $ with the ideal clean model $f_{\mathrm{clean}}$ on the
test benchmark $D_{\mathrm{test}}$.
We write the performance gap $\Delta_{\mathrm{perf}}$ as:
\begin{equation*}
\small
\begin{split}
\Delta_{\mathrm{perf}}
\;:=\;
\bigl|&\mathrm{Perf}(M (f_{\mathrm{con}}), D_{\mathrm{test}}) 
-\mathrm{Perf}(f_{\mathrm{clean}}, D_{\mathrm{test}})\bigr|.
\end{split}
\end{equation*}
Directly minimizing $\Delta_{\mathrm{perf}}$ is intractable because standard
metrics (e.g., multiple-choice accuracy) are discrete and non-differentiable.
We therefore adopt a differentiable, distribution-level surrogate.
For each $x\in D_{\mathrm{test}}$, let $p_{\mathrm{con}}(\cdot\mid x)$ and
$p_{\mathrm{clean}}(\cdot\mid x)$ denote the output distributions of
$M(f_{\mathrm{con}})$ and $f_{\mathrm{clean}}$, respectively.
By viewing performance as $\mathbb{E}{y\sim p(\cdot\mid x)}[u(x,y)]$ for a
bounded utility $u(x,y)\in[0,1]$, the performance gap can be related to a
distance between $p_{\mathrm{con}}$ and $p_{\mathrm{clean}}$, requiring only
boundedness and thus covering a range of evaluation metrics.



 Theorem in Appendix ~\ref{app:kl-bound} shows that the average KL divergence between the mitigated contaminated model and the clean model upper-bounds the performance gap. 
We therefore use KL minimization as a principled surrogate objective. 
This bound is only used as motivation and does not guarantee our algorithm's behavior. 
Next, we instantiate $M$ so that the KL objective can be optimized efficiently.

\subsection{Embedding-Space Perturbation for Inference-Time Mitigation}
\label{subsec:embedding}

We implement the mitigation mechanism $M$ by applying small, controlled
perturbations in the embedding space, while keeping the discrete input text
unchanged.
This design preserves semantic and difficulty invariance of benchmark
items, which input-space transformations such as paraphrasing or rewriting may
violate.

Formally, let $e(x)\in\mathbb{R}^{L\times d}$ denote the input embedding sequence
for a benchmark question $x$.
We define the mitigated prediction as
\begin{equation}
\small
M(f)(x) = f(e(x) + \delta(x)),
\quad
\|\delta(x)\|_{\infty} \le \zeta,
\end{equation}
where $\delta(x)$ is a bounded perturbation with a small budget $\zeta$.
Constraining $\delta(x)$ in an $\ell_{\infty}$-ball encourage that the perturbed
embedding remains in a local neighborhood of $e(x)$, aiming to preserve the
underlying semantics and difficulty while disrupting memorized surface patterns.

This reasoning leads to the ideal per-sample objective of aligning the perturbed
contaminated model with clean behavior:
\begin{equation}
\small
\begin{aligned}[b]
\min_{\{\delta(x)\}_{x\in D}}& \quad 
\frac{1}{|D|}
\sum_{x\in D}
\mathrm{KL}\big(
f_{\mathrm{con}}(e(x)+\delta(x))
\,\big\|\,
f_{\mathrm{clean}}(e(x))
\big) \\
&\text{s.t.} \quad \|\delta(x)\|_{\infty} \le \zeta,\;\forall x .
\end{aligned}
\end{equation}
Directly optimizing a separate perturbation for each input is computationally
infeasible at evaluation time.
We therefore amortize the optimization by learning a generator $G_{\theta}$. After training, we can use such a generator to generate $\delta(x)$ for all
$x \in D_{\mathrm{test}}$ (and other unseen inputs). Our amortized objective
becomes:
\begin{equation}
\small
\label{eq:amortized_obj}
\begin{split}
\min_{\theta} \;\; & \frac{1}{|D|}\sum_{x\in D}
\mathrm{KL}\Big(
f_{\mathrm{con}}\big(e(x) + G_{\theta}(e(x))\big) 
\,\big\|\,
f_{\mathrm{clean}}\big(e(x)\big)
\Big) \\
&\text{s.t.} \;\;  \big\| G_{\theta}(e(x)) \big\|_{\infty} \le \zeta, \quad \forall x\in D .
\end{split}
\end{equation}
where $D$ is an auxiliary dataset used to train $G_{\theta}$ sampled from evaluated benchmark.
\subsection{Generator Design}
\label{subsec:generator-impl}

\paragraph{Setup}
We assume access to a contaminated model $f_{\mathrm{con}}$, a reference model
$f_{\mathrm{ref}}$, and a small auxiliary dataset $D_{\mathrm{aux}}$ sampled from
the benchmark.
The reference model serves as a practical proxy for clean behavior during
generator training.
Since a strictly clean model $f_{\mathrm{clean}}$ is often unavailable,
$f_{\mathrm{ref}}$ is typically chosen as the same-architecture base model (or an
earlier checkpoint) from which $f_{\mathrm{con}}$ is fine-tuned, and is expected
to have less benchmark-specific exposure.
This choice is standard in contamination-related analyses
\citep{zhu2025establishingtrustworthyllmevaluation,Nicholas2021referencemodel,carlini2022membership}.

\paragraph{Generator Architecture}
We parameterize the perturbation generator as a lightweight sequence-to-sequence
network.
Let ${G}_{\theta}$ denote an  generator that maps the input
embedding $e(x)\in\mathbb{R}^{L\times d}$ to a raw perturbation. To enforce the perturbation budget, we apply a scaled $\tanh$ transformation:
\[
\delta(x) = \zeta \cdot \tanh(G_{\theta}(x)),
\]
which guarantees $\|\delta(x)\|_{\infty} \le \zeta$ by construction.
The perturbed embedding $e(x)+\delta(x)$ is then fed to $f_{\mathrm{con}}$ at
inference time.
Unless otherwise specified, ${G}_{\theta}$ is instantiated as a small
decoder-only Transformer \citep{vaswani2023attentionneed}; architectural details are provided in
Appendix~\ref{app:generator-arch}.

\paragraph{Training Objective}
For a training sample $x \sim D_{\mathrm{aux}}$, let
$p^{\mathrm{con}} = f_{\mathrm{con}}(e(x)+\delta(x))$ and
$p^{\mathrm{ref}} = f_{\mathrm{ref}}(e(x))$ denote the output distributions of
the contaminated and reference models, respectively.
We train the generator using a composite KL+CE objective:
\begin{equation}
\label{eq:total_loss}
\begin{split}
\mathcal{L}(\theta) = 
\mathbb{E}_{x \sim D_{\mathrm{aux}}} \Big[
&\lambda_{\mathrm{KL}} \cdot 
\mathrm{KL}\big(p^{\mathrm{con}} \,\|\, p^{\mathrm{ref}}\big) \\
+ \;&\lambda_{\mathrm{CE}} \cdot 
\mathrm{CE}\big(p^{\mathrm{con}}, y_{\mathrm{ref}}\big)
\Big],
\end{split}
\end{equation}
where $y_{\mathrm{ref}}$ denotes hard token labels sampled from
$f_{\mathrm{ref}}$, and $\lambda_{\mathrm{KL}}, \lambda_{\mathrm{CE}}$ control the
relative weights of the two terms.
The KL term directly optimizes the surrogate objective motivated by
Theorem~\ref{thm:kl-bound}, while the cross-entropy term stabilizes training and
prevents degenerate solutions. In the appendix, Algorithm~\ref{alg:train_gen} summarizes the training procedure.

\section{Experiment}

\subsection{Experiment Setup}


\paragraph{Dataset}
We evaluate on two benchmarks: MMLU~\citep{hendryckstest2021} and TruthfulQA~\citep{lin2022truthfulqameasuringmodelsmimic}.
MMLU is a large-scale multiple-choice benchmark spanning diverse academic and professional domains, while TruthfulQA measures truthfulness and resistance to common misconceptions.
In our experiments, these benchmarks serve dual roles. First, they define the evaluation benchmarks. Second, to simulate benchmark contamination, a small subset of benchmark items is intentionally allowed to leak into the training data when constructing contaminated models. All benchmark items used for contamination are drawn exclusively from the corresponding benchmark (MMLU or TruthfulQA). We also test the proposal on the code generation task, and shown in the Appendix \ref{app:codegen}.

\paragraph{Models} We conduct experiments on three instruction-tuned large language models : LLaMA-3-8B-Instruct \cite{grattafiori2024llama3herdmodels}, Qwen Family \cite{qwen2025qwen25technicalreport}, and Mistral-7B-Instruct-v0.3 \cite{jiang2023mistral7b}. These models represent different model families, allowing us to evaluate the generality of our decontamination method across heterogeneous architectures. Before simulating contamination, we start from the corresponding pretrained (unfine-tuned) model for each architecture. This pretrained model serves as the common initialization for all experiments and is also used as the reference model in our decontamination framework.
Fine-tuning is then performed either with or without benchmark leakage, depending on the contamination setting. These model families span heterogeneous architectures, allowing us to evaluate the generality of our decontamination method.

\begin{table*}[hbt]
\centering
\scriptsize
\setlength{\tabcolsep}{3.2pt}
\renewcommand{\arraystretch}{1.15}
\begin{tabular}{@{}lllc ccccc ccc@{}}
\toprule
\multirow{2}{*}{Model} & \multirow{2}{*}{Split} & \multirow{2}{*}{Method} & \multirow{2}{*}{Access} &
\multicolumn{3}{c}{TruthfulQA ($o=3$)} & \multicolumn{3}{c}{MMLU ($o=3$)} \\
\cmidrule(lr){5-7}\cmidrule(lr){8-10}
 &  &  &  & Exact & Semantic-level & Domain-level & Exact & Semantic-level & Domain-level \\
\midrule

\multirow{7}{*}{Mistral}
 & $f_{\mathrm{clean}}$ & baseline & -- &
0.287 & 0.249 & 0.269 & 0.451 & 0.434 & 0.406 \\
\cmidrule(lr){2-10}
 & \multirow{6}{*}{$f_{\mathrm{con}}$} & W/O Decontamination & -- &
0.976 (0.689) & 0.971 (0.722) & 0.861 (0.592) & 0.892 (0.441) & 0.875 (0.441) & 0.475 (0.069) \\
 &  & TED & Black-box &
\underline{0.535 (0.248)} & \underline{0.579 (0.330)} & 0.740 (0.471) & 0.224 (0.227) & \underline{0.388 (0.046)} & \underline{0.375 (0.031)} \\
 &  & ITD & Black-box &
0.976 (0.689) & 0.971 (0.722) & 0.861 (0.592) & 0.886 (0.435) & 0.863 (0.429) & 0.471 (0.065) \\
 &  & Shortcut Neuron & White-box &
0.668 (0.381) & 0.634 (0.385) & \underline{0.636 (0.367)} & \underline{0.380 (0.071)} & 0.386 (0.048) & 0.239 (0.167) \\
 &  & Short Circuit & White-box &
0.933 (0.646) & 0.914 (0.665) & 0.793 (0.524) & 0.833 (0.382) & 0.808 (0.374) & 0.478 (0.072) \\
 &  & DeconIEP(Ours) & White-box &
\textbf{0.533 (0.246)} & \textbf{0.551 (0.302)} & \textbf{0.565 (0.296)} & \textbf{0.458 (0.007)} & \textbf{0.406 (0.028)} & \textbf{0.417 (0.011)} \\
\midrule

\multirow{7}{*}{Qwen2.5}
 & $f_{\mathrm{clean}}$ & baseline & -- &
0.555 & 0.531 & 0.572 & 0.670 & 0.671 & 0.673 \\
\cmidrule(lr){2-10}
 & \multirow{6}{*}{$f_{\mathrm{con}}$} & W/O Decontamination & -- &
0.976 (0.421) & 0.952 (0.421) & 0.885 (0.313) & 0.905 (0.235) & 0.872 (0.201) & 0.744 (0.071) \\
 &  & TED & Black-box &
0.880 (0.325) & 0.900 (0.369) & 0.846 (0.274) & 0.837 (0.167) & 0.882 (0.211) & 0.713 (0.040) \\
 &  & ITD & Black-box &
0.976 (0.421) & 0.947 (0.416) & 0.899 (0.327) & 0.898 (0.228) & 0.868 (0.197) & 0.751 (0.078) \\
 &  & Shortcut Neuron & White-box &
0.684 (0.129) & \textbf{0.526 (0.005)} & 0.394 (0.178) & 0.416 (0.254) & 0.420 (0.251) & 0.371 (0.302) \\
 &  & Short Circuit & White-box &
\underline{0.650 (0.095)} & \underline{0.622 (0.091)} & \textbf{0.614 (0.042)} & \underline{0.807 (0.137)} & \underline{0.757 (0.086)} & \underline{0.640 (0.033)} \\
 &  & DeconIEP(Ours) & White-box &
\textbf{0.620 (0.065)} & 0.632 (0.101) & \underline{0.641 (0.069)} & \textbf{0.785 (0.115)} & \textbf{0.632 (0.039)} & \textbf{0.646 (0.027)} \\
\midrule

\multirow{7}{*}{LLaMA-3}
 & $f_{\mathrm{clean}}$ & baseline & -- &
0.474 & 0.467 & 0.452 & 0.576 & 0.565 & 0.549 \\
\cmidrule(lr){2-10}
 & \multirow{6}{*}{$f_{\mathrm{con}}$} & W/O Decontamination & -- &
0.986 (0.512) & 0.981 (0.514) & 0.928 (0.476) & 0.921 (0.345) & 0.901 (0.336) & 0.669 (0.120) \\
 &  & TED & Black-box &
0.966 (0.493) & 0.951 (0.485) & 0.923 (0.471) & 0.864 (0.288) & 0.842 (0.277) & 0.638 (0.089) \\
 &  & ITD & Black-box &
0.943 (0.469) & 0.957 (0.490) & 0.899 (0.447) & 0.886 (0.310) & 0.844 (0.279) & 0.648 (0.098) \\
 &  & Shortcut Neuron & White-box &
{0.818 (0.345)} & 0.852 (0.385) & 0.726 (0.274) & \textbf{0.690 (0.114)} & 0.682 (0.117) & \underline{0.482 (0.068)} \\
 &  & Short Circuit & White-box &
\underline{0.593 (0.120)} & \underline{0.608 (0.141)} & \underline{0.625 (0.173)} & \underline{0.701 (0.125)} & \underline{0.680 (0.115)} & 0.707 (0.157) \\
 &  & DeconIEP(Ours) & White-box &
\textbf{0.536 (0.062)} & \textbf{0.519 (0.053)} & \textbf{0.553 (0.101)} & 0.741 (0.165) & \textbf{0.537 (0.028)} & \textbf{0.536 (0.014)} \\

\bottomrule
\end{tabular}
\caption{Decontamination results for $o=3$ across different models. Values denote accuracy, with parentheses showing RC (smaller is better). Best RC is in bold and second best is underlined within each model and split. ($\zeta=10^{-3}$)}
\label{tab:decontamination_o3_all}
\end{table*}


\begin{figure}[t]
  \centering

  \begin{subfigure}[t]{\linewidth}
    \centering
    \includegraphics[width=0.9\linewidth]{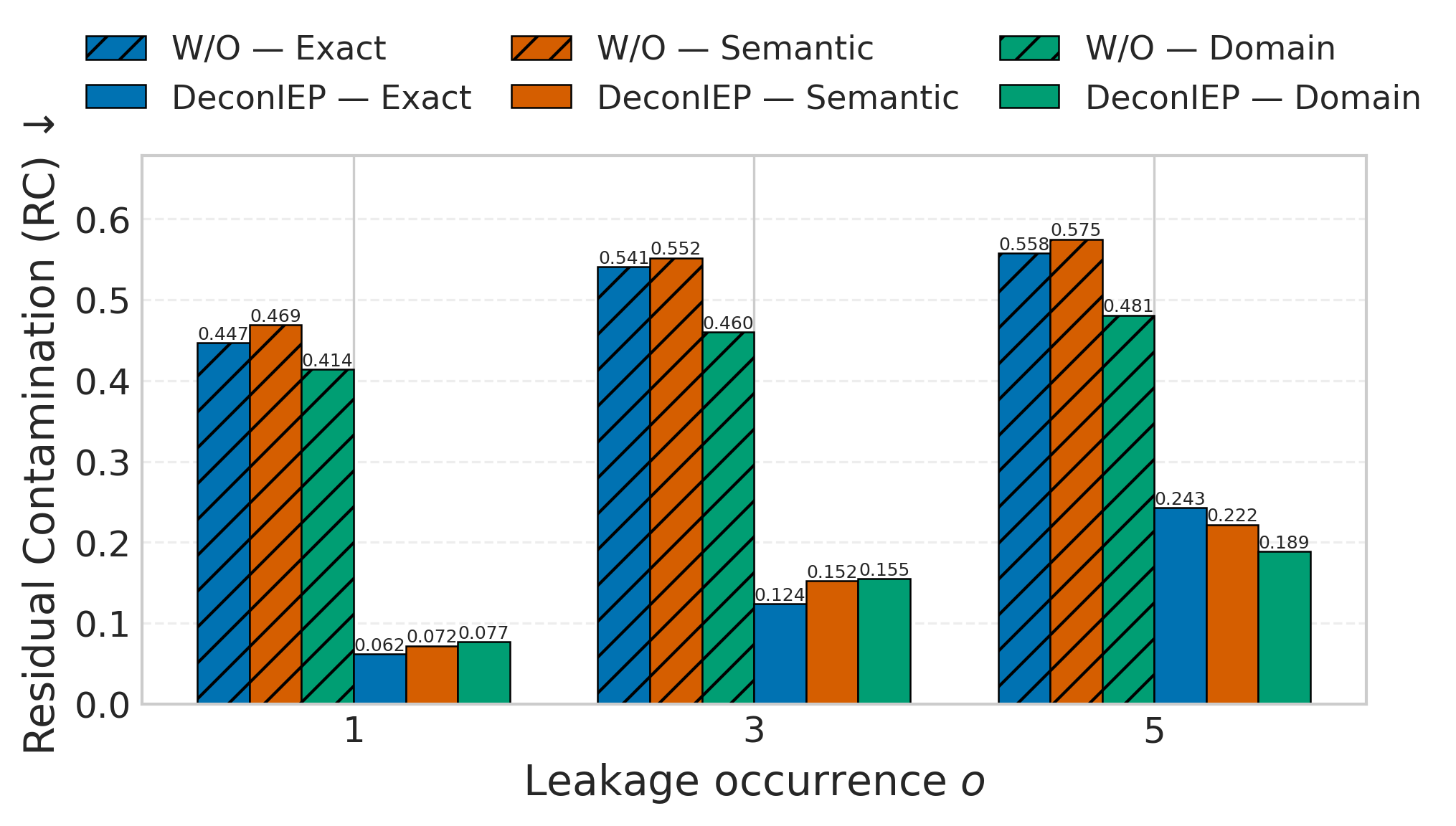}
    \caption{TruthfulQA.}
    \label{fig:rc_bars_tqa}
  \end{subfigure}

  \vspace{0.6em}

  \begin{subfigure}[t]{\linewidth}
    \centering
    \includegraphics[width=0.9\linewidth]{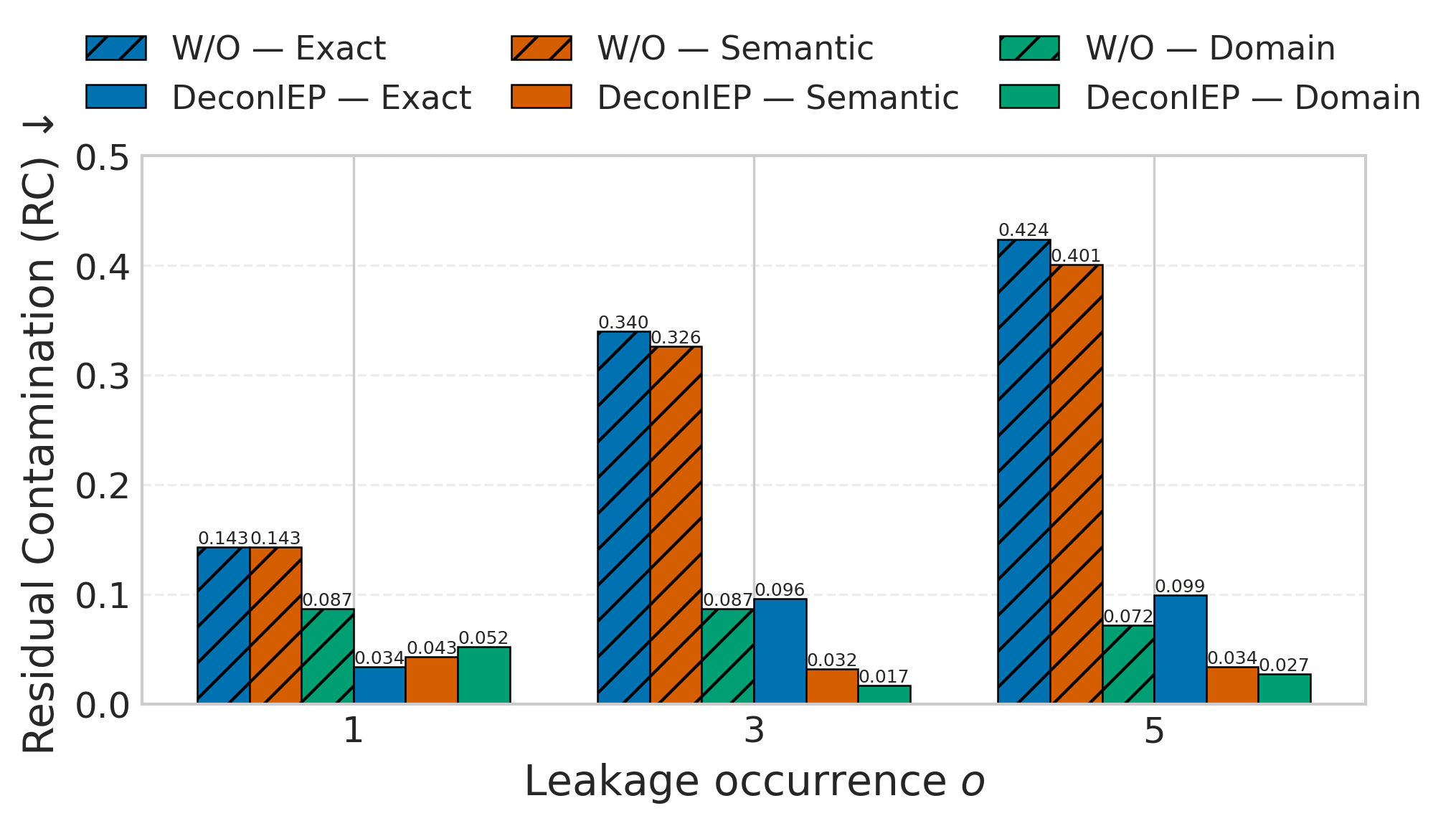}
    \caption{MMLU.}
    \label{fig:rc_bars_mmlu}
  \end{subfigure}

  \caption{Average Residual Contamination (RC) of all models across leakage occurrences $o$ under different contamination levels (lower is better).( $\zeta=10^{-3}$)}
  \label{fig:rc_bars}
\end{figure}

\paragraph{Contamination simulation and evaluation}
Following common practice in prior work~\citep{zhu2025establishingtrustworthyllmevaluation}, we simulate benchmark contamination in a controlled manner.
For each benchmark, we construct a contaminated model \( f_{\mathrm{con}} \) by fine-tuning the pretrained model on a mixture of:
(i) a general instruction-tuning dataset OpenOrca, denoted as \( D_{\mathrm{train}}^{\mathrm{clean}} \), and
(ii) a small set of leaked benchmark items \( D_{\mathrm{con}}  \), drawn from either MMLU or TruthfulQA.
Each benchmark item in \( D_{\mathrm{con}} \) is repeatedly included in training \( o \in \{1,3,5\} \) times to simulate different contamination strengths.
To isolate the effect of benchmark leakage, we also train a clean counterpart \( f_{\mathrm{clean}} \) by fine-tuning on OpenOrca only, using the same total number of training samples as \( f_{\mathrm{con}} \), but without including any benchmark items.

We evaluate decontamination performance under three contamination settings, each associated with a different construction of the test set \( D_{\mathrm{test}} \):
(i) \emph{Exact contamination:} \( D_{\mathrm{test}} \) consists of a subset of the leaked benchmark items from \( D_{\mathrm{con}} \), evaluating direct memorization effects;
(ii) \emph{Semantic-level contamination:} \( D_{\mathrm{test}} \) consists of paraphrases of the leaked benchmark items, which are semantically equivalent but not observed during training, with ground-truth answers unchanged;
and
(iii) \emph{Domain-level contamination:} \( D_{\mathrm{test}} \) is sampled from the same benchmark distribution but is strictly disjoint from \( D_{\mathrm{con}} \), evaluating same-domain generalization rather than direct leakage.
For benign (clean) utility evaluation, we additionally evaluate on an external benchmark that is not used to construct \( D_{\mathrm{con}} \).
Specifically, when contamination is simulated on MMLU, TruthfulQA is used as the clean test benchmark, and vice versa.

\paragraph{Implementation Details}
The noise generator \( G_{\theta} \) is a 4-layer decoder-only Transformer (\( n = 4 \)).
The generator is trained on the same benchmark used to construct the contamination set (e.g., MMLU for MMLU-contaminated models), using benchmark samples that are disjoint from the test set in each evaluation setting.
We set the perturbation budget \( \zeta = 10^{-3} \), learning rate \( lr = 1 \times 10^{-5} \), dropout rate to 0.2, and use an auxiliary dataset of size \( |D_{\mathrm{aux}}| = 400 \), which is sampled from contaminated benchmark and disjoint from \( D_{\mathrm{test}} \).
Additional details shown in Appendix~\ref{app:more experimrnt}.

\paragraph{Metric}

We evaluate decontamination using two metrics:
Residual Contamination (RC) and
Benign Utility Drop (BUD).
RC measures how close the decontaminated model $M(f_{\mathrm{con}})$
is to the uncontaminated model $f_{\mathrm{clean}}$ on contaminated data,
while BUD measures utility loss on clean data. 

Smaller values indicate better decontamination. Formal definition see Appendix\ref{app: Definition of RC and BUD}.

\paragraph{Comparison methods}
We compare our method with representative inference-time decontamination
baselines explained in Section \ref{sec:related-work}: TED and ITD as
black-box methods,  Shortcut Neuron  and Short Circuit 
as white-box methods.

\begin{figure}[t]
    \centering
    \begin{subfigure}[t]{0.95\linewidth}
        \centering
        \includegraphics[width=\linewidth]{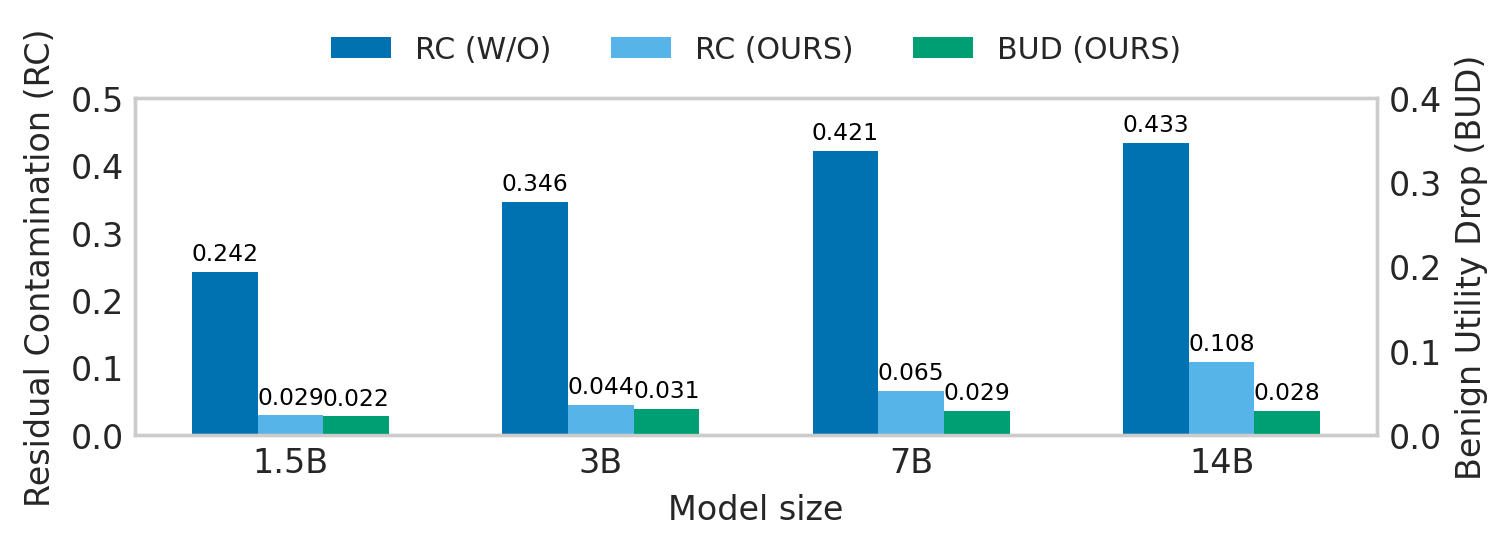}
        \caption{TruthfulQA ($o=3$).}
        \label{fig:scale_truthfulqa}
    \end{subfigure}

    \vspace{0.35em}

    \begin{subfigure}[t]{0.95\linewidth}
        \centering
        \includegraphics[width=\linewidth]{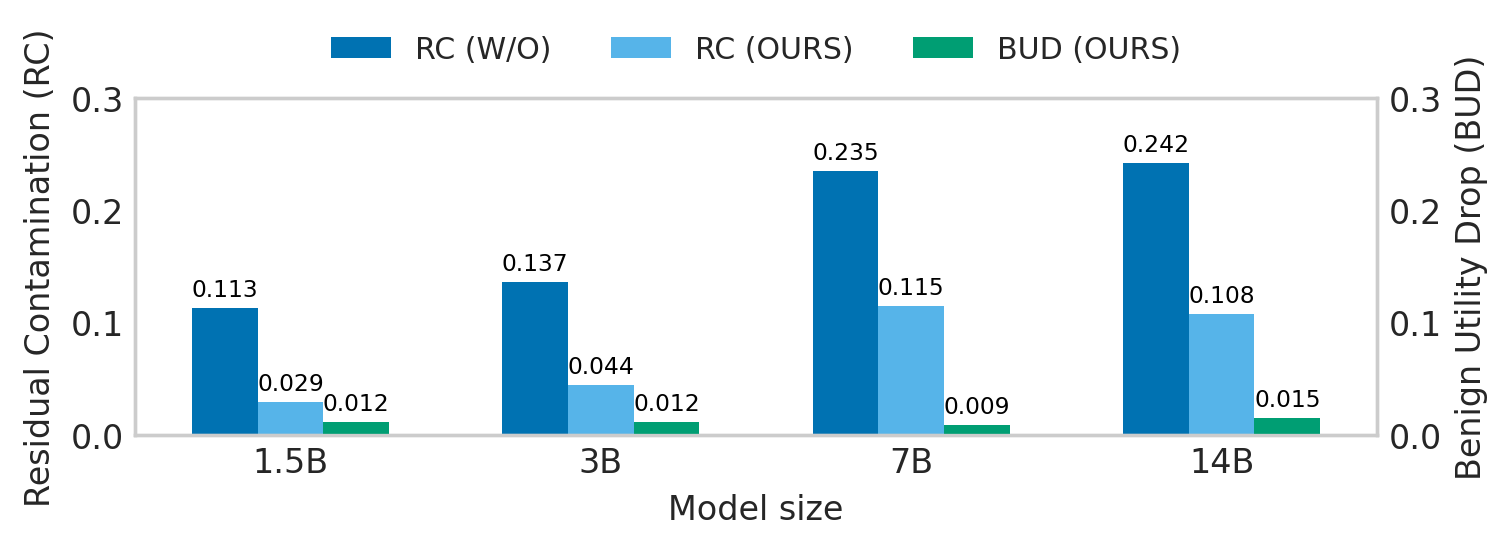}
        \caption{MMLU ($o=3$).}
        \label{fig:scale_tqa}
    \end{subfigure}

    \caption{{Scaling behavior on Qwen2.5-Instruct models ($o=3$).}
    We compare RC on exact domian and BUD across model sizes (1.5B/3B/7B/14B) on (a) TruthfulQA and (b) MMLU.
    Our method remains effective and stable under scaling with a fixed embedding perturbation budget $\zeta=10^{-3}$.}
    \label{Fig:scale_results}
\end{figure}

\subsection{Evaluation of Decontamination Effectiveness}

We first evaluate decontamination by whether the performance of the post-mitigation model approaches that of the uncontaminated model $f_{\mathrm{clean}}$. Table ~\ref{tab:decontamination_o3_all} reports accuracy and Residual
Contamination (RC; lower is better) on TruthfulQA and MMLU under $o=3$, $o=1$ and $o=5$ see Appendix \ref{app:The Full Result}.
Overall, DeconIEP achieves the lowest or near-lowest RC across models,
benchmarks, and contamination granularities, and remains robust under three-level contamination. While some baselines are best
in isolated cases, their improvements are not consistent. 

Figure~\ref{fig:rc_bars} further varies the occurrence level $o$.
Although larger $o$ increases residuals for the unmitigated model, {DeconIEP}
consistently reduces RC and keeps accuracy closer to $f_{\mathrm{clean}}$. To assess whether decontamination introduces unintended degradation on clean tasks,
we evaluate \emph{benign utility } using a cross-benchmark clean evaluation.
Specifically, to isolate benign utility from contamination-related effects,
models contaminated on MMLU are evaluated on the unrelated TruthfulQA benchmark,
and vice versa. As shown in Table~\ref{table:BUD_results}, {DeconIEP} incurs minimal BUD across all model families, whereas several white-box baselines
exhibit substantially larger utility degradation.
These results indicate that the proposed embedding-space intervention effectively
mitigates contamination while preserving benign performance, with advantages that
remain stable across heterogeneous architectures.

\begin{table}[bht]
\centering
\scriptsize
\setlength{\tabcolsep}{4pt}
\renewcommand{\arraystretch}{1.15}
\begin{tabular}{@{}lllcc@{}}
\toprule
\multirow{2}{*}{Model} & \multirow{2}{*}{Method} & \multirow{2}{*}{Access} & \multicolumn{2}{c}{Benign Utility Drop $\downarrow$} \\
\cmidrule(lr){4-5}
 &  &  & TruthfulQA & MMLU \\
\midrule

\multirow{5}{*}{Mistral}
 & DeconIEP        & White-box & 0.019 & 0.016 \\
 & TED             & Black-box & 0.028 & 0.030 \\
 & ITD             & Black-box & 0.000 & 0.001 \\
 & Shortcut Neuron & White-box & 0.129 & 0.152 \\
 & Short Circuit   & White-box & 0.057 & 0.077 \\
\midrule

\multirow{5}{*}{Qwen2.5}
 & DeconIEP        & White-box & 0.009 & 0.029 \\
 & TED             & Black-box & 0.110 & 0.022 \\
 & ITD             & Black-box & 0.000 & 0.002 \\
 & Shortcut Neuron & White-box & 0.207 & 0.256 \\
 & Short Circuit   & White-box & 0.172 & 0.238 \\
\midrule

\multirow{5}{*}{LLaMA-3}
 & DeconIEP        & White-box & 0.031 & 0.041 \\
 & TED             & Black-box & 0.012 & 0.035 \\
 & ITD             & Black-box & 0.010 & 0.016 \\
 & Shortcut Neuron & White-box & 0.057 & 0.190 \\
 & Short Circuit   & White-box & 0.105 & 0.161 \\
\bottomrule
\end{tabular}

\caption{{BUD under cross-benchmark evaluation ($o=3$).}
We contaminate each model on MMLU and measure the utility drop on TruthfulQA to assess benign performance on clean data, and vice versa. ($\zeta=10^{-3}$)}
\label{table:BUD_results}
\end{table}

\subsection{Evaluation across Model scale}
We evaluate our decontamination method across model scales using four Qwen2.5
models \citep{qwen2025qwen25technicalreport}: Qwen2.5-1.5B-Instruct,
Qwen2.5-3B-Instruct, Qwen2.5-7B-Instruct, and Qwen2.5-14B-Instruct.
Figure~\ref{Fig:scale_results} shows that our method maintains stable
decontamination effectiveness across all scales, indicating that it generalizes
beyond a specific model size, with only mild variation across model capacities.

\subsection{Semantic preservation}
\begin{figure}[t]
  \centering
  \includegraphics[width=1\linewidth]{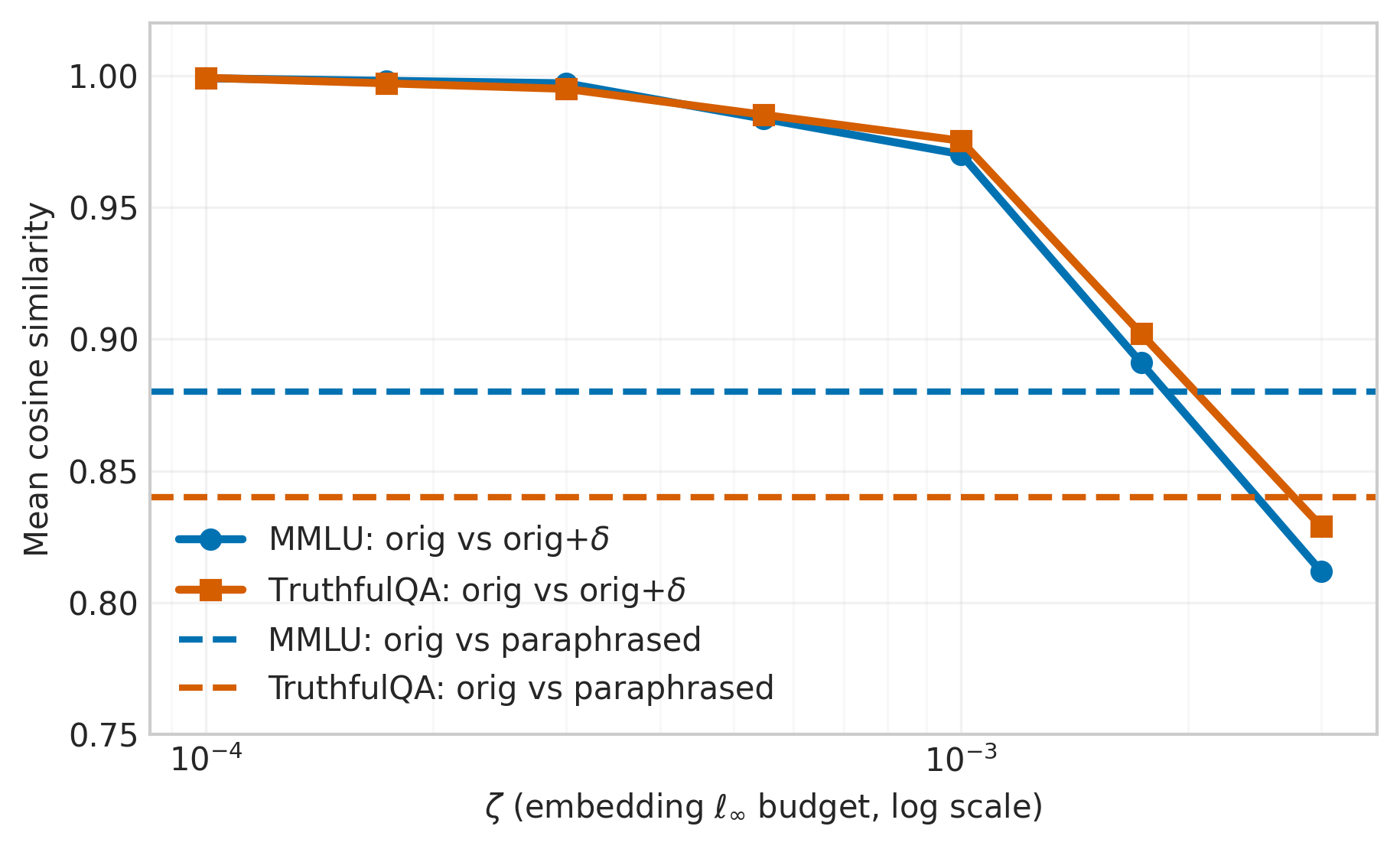}
  \caption{{Semantic invariance under embedding perturbations.}
  Mean cosine similarity between original and perturbed embeddings (orig vs.\ orig+$\delta$)
  versus $\zeta$ on MMLU and TruthfulQA. Dashed lines show orig vs.\ paraphrased
  as a semantic-preserving reference.}
  \label{fig:cosine-sweep}
\end{figure}

Then, we verify that embedding perturbations preserve input semantics.
We measure the average cosine similarity between the original embedding and the
perturbed embedding,
\[
\mathrm{Cos}(\zeta)
\;=\;
\mathbb{E}_{x}\Big[\cos\big(e(x),\,e(x)+\delta(x)\big)\Big],
\]
using the same pooling operator as in our implementation.
Figure~\ref{fig:cosine-sweep} shows that for practical budgets $\zeta\le10^{-3}$,
cosine similarity remains close to $1$, and consistently exceeds the similarity
between the original prompt and its rewritten counterpart. This suggests that
embedding-level intervention introduces less semantic drift than rewriting while
still enabling effective decontamination.

\subsection{Ablation Study}
\label{subsec:ablation}


\begin{table}[t]
\centering
\small
\setlength{\tabcolsep}{5.5pt}
\renewcommand{\arraystretch}{1.15}
\begin{tabular}{@{}lccc|ccc@{}}
\toprule
\multirow{2}{*}{  $\zeta$} &
\multicolumn{3}{c}{TruthfulQA} &
\multicolumn{3}{c}{MMLU} \\
\cmidrule(lr){2-4}\cmidrule(lr){5-7}
& RC $\downarrow$ & BUD $\downarrow$ &  &
RC $\downarrow$ & BUD $\downarrow$ &  \\
\midrule
$1\mathrm{e}{-5}$  & 0.512 & 0.000 &  & 0.344 & 0.000 &  \\
$3\mathrm{e}{-5}$  & 0.393 & 0.009 &  & 0.304 & 0.000 &  \\
$1\mathrm{e}{-4}$  & 0.226 & 0.024 &  & 0.214 & 0.022 &  \\
$3\mathrm{e}{-4}$  & 0.096 & 0.034 &  & 0.184 & 0.031 &  \\
$1\mathrm{e}{-3}$  & 0.062 & 0.038 &  & 0.165 & 0.041 &  \\
$3\mathrm{e}{-3}$  & 0.024 & 0.050 &  & 0.034 & 0.070 &  \\
$1\mathrm{e}{-2}$  & 0.022 & 0.062 &  & 0.032 & 0.102 &  \\
\bottomrule
\end{tabular}
\caption{{RC--BUD trade-off induced by the perturbation budget $\zeta$.}
We report RC and BUD for DeconIEP
under different $\zeta$ on TruthfulQA and MMLU for contaminated Llama 3 model, $o=3$ .}
\label{tab:pareto_three_line}
\end{table}

\paragraph{Controllability via the perturbation budget $\zeta$}
We next treat $\zeta$ as an explicit control knob that trades off
decontamination strength against benign utility. Table~\ref{tab:pareto_three_line}
shows the relationship between RC and BUD induced by varying $\zeta$.
As $\zeta$ increases, DeconIEP moves toward lower RC at the cost of higher BUD,
forming a clear and monotonic trade-off curve. Compared with representative
baselines (see Appendix \ref{app:pareto}), DeconIEP provides a more favorable RC--BUD region, and the practical
budget regime yields strong contamination suppression with limited utility loss.

\paragraph{Effect of reference model}
Finally, we examine the role of the reference model. While our method uses a
reference model to guide perturbation generation, a perfectly clean anchor may
be unavailable in practice. We therefore test robustness when the reference
model is mildly contaminated.

\begin{figure}[t]
  \centering
  \includegraphics[width=1\linewidth]{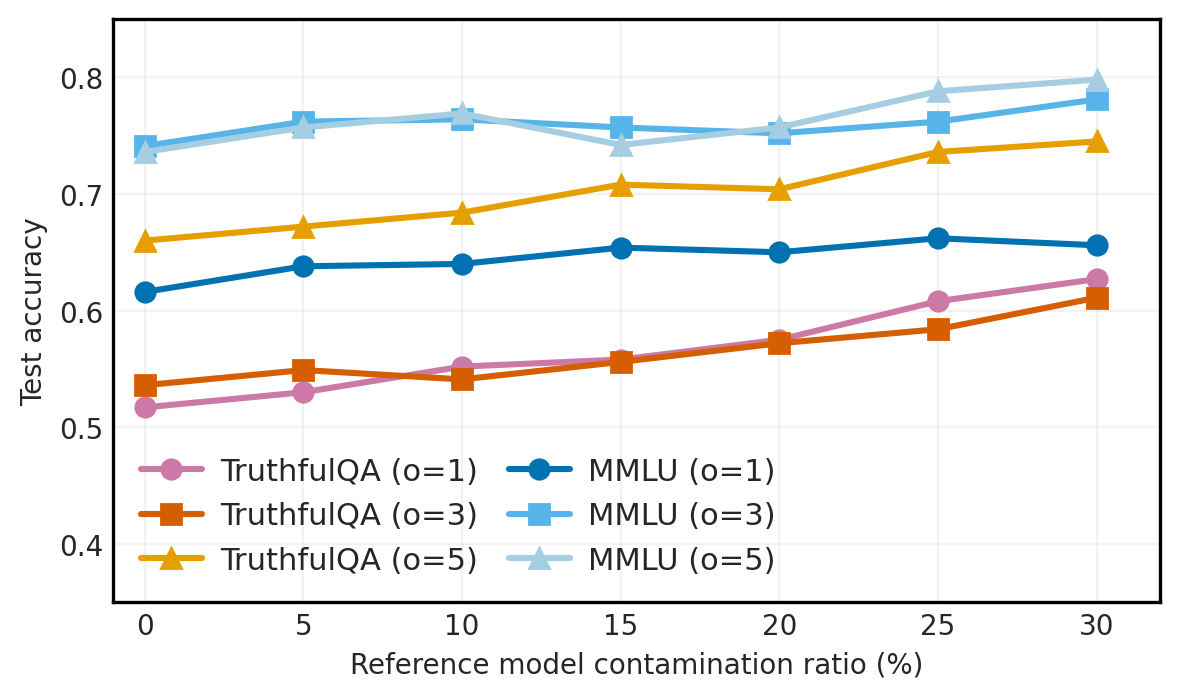}
  \caption{Test accuracy under increasing contamination in the reference model.}
  \label{fig:ref-model-sensitivity}
\end{figure}

We construct imperfect references by fine-tuning the reference model on data
containing different proportions of benchmark test items, yielding varying
contamination ratios, and then run our method under the same contaminated
evaluation setting.
Figure~\ref{fig:ref-model-sensitivity} shows that performance is largely stable
as the reference contamination increases from $0\%$ to $30\%$ across benchmarks
and occurrence levels. This indicates that the reference need not be strictly
clean: even mildly contaminated, it provides a sufficiently informative anchor
distribution to steer the evaluated model away from contamination-driven
behavior. In practice, a readily available same-architecture base model can
serve as a  reference.

We also do experiments to investigate the difference between our perturbation and random noise, how the sample size of $D_{\mathrm{aux}}$ affects our proposal; due to space constraints, we report these results in Appendix~\ref{app:randomnoise} and \ref{app:auxsize}.


\section{Conclusion}

Test-set contamination inflates benchmark scores, and detect-then-filter evaluation
fails under moderate contamination due to detector errors. We propose {DeconIEP},
an inference-time method that applies small, $\ell_\infty$-bounded perturbations to
input embeddings to align a contaminated model with a less-contaminated
reference. Across multiple open-weight LLM families and two benchmarks, DeconIEP reduces
residual contamination with minimal benign utility drop.

\section{Limitations}
Our study has several limitations. First, DeconIEP is a white-box method: it requires access to input embeddings (and, in our training setup, gradients) to generate and apply perturbations, which limits applicability to closed-model APIs without additional approximations. Second, DeconIEP relies on a reference model to provide a comparatively less-contaminated behavioral anchor. While we empirically observe robustness to moderate reference contamination, performance may degrade when the reference is heavily contaminated, distributionally mismatched, or differs substantially in alignment behavior from the evaluated model. Then, although our intervention is bounded and empirically exhibits strong semantic invariance (e.g., high cosine similarity under small $\zeta$), we do not provide formal guarantees that semantics/difficulty are preserved for all inputs. Finally, DeconIEP introduces additional inference overhead for perturbation generation and reference-guided objectives; while lightweight relative to multi-query black-box baselines, reducing overhead for very large-scale evaluation remains an important direction.
\bibliography{custom}

\appendix

\section{Related Work}
\label{app:related-work}

\subsection{Data Contamination}
Test-set leakage in widely used benchmarks (e.g., MMLU, GSM8K) has become pervasive, where inflated scores may reflect memorization rather than genuine generalization \citep{chen2025benchmarking,deng2024investigating,xu2024benchmarking}.
When an auditor has full access to the training corpus, contamination can be identified by direct overlap checks (exact match or $n$-gram similarity) between training documents and benchmark instances \citep{brown2020language,gao2021framework}.
In recent years, the distribution of open-source models via platforms like Hugging Face has become widespread, but information disclosure regarding their training data remains scarce, leaving no means to verify whether benchmarking is being conducted properly. A straightforward approach is to use MIA to detect whether test instances were seen during training and filter out detected contaminated items before evaluation. However, imperfect detection can leave residual contamination and bias the reported scores.
 This motivate \emph{inference-time} approaches that aim to correct contaminated behavior without modifying model weights or relying on the training data.
\subsection{Inference-time Decontamination}
\paragraph{Black-box methods}
Early inference-time decontamination methods assume evaluators only have API access to the model, without access to internal parameters or activations.
As summarized in Table~\ref{tab:comparison}, black-box approaches mainly rely on output-side interventions.
TED~\citep{dong2024generalization} frames decontamination as distribution calibration: it keeps model parameters fixed and approximates a ``clean'' output distribution via repeated stochastic decoding.
ITD~\citep{zhu2024inference} rewrites potentially contaminated prompts using an auxiliary model like GPT , aiming to reduce direct reuse of memorized patterns.
These methods are convenient to deploy, but they typically incur high inference cost  including extra model calls (e.g., auxiliary LLM rewriting), repeated decoding/sampling and can be limited against strongly memorized contamination; moreover, rewriting-based methods do not strictly preserve semantics and may change question difficulty ~\citep{xu2024benchmarking}.

\paragraph{White-box methods}
For benchmark evaluation, white-box access has become increasingly practical due to the availability of open-weight LLMs.
White-box access enables targeted inference-time interventions using internal signals such as parameters, intermediate activations, and layer-wise representations.
Recent work in this regime can be grouped into three directions (Table~\ref{tab:comparison}).

\emph{(i) Neuron/activation patching.}
This line localizes Neurons in LLM that drive contaminated behavior and patches their activations during inference.
Shortcut Neuron Analysis~\citep{zhu2025establishingtrustworthyllmevaluation} uses a reference model to identify such shortcut neurons and then edits their activations at test time.
As shown in Table~\ref{tab:comparison}, it can achieve high decontamination efficacy, but often leads to a large clean-evaluation drop when the edited neurons overlap with general-purpose computation.

\emph{(ii) Architectural bypassing to weaken memorization.}
A second line modifies specific modules in the forward pass to reduce memorization-driven behavior during decoding.
Since contamination effects often manifest as answer retrieval on leaked items~\citep{xu2024benchmarking}, weakening memorization at inference time can reduce contamination-induced score inflation.
Menta et al.~\citep{menta2025analyzing} propose attention short-circuiting by replacing attention mixing with an identity operation, so that values are forwarded without cross-token attention.
They show that bypassing attention in deeper layers can substantially reduce memorized content generation.
However, as summarized in Table~\ref{tab:comparison}, this approach will also cause a large clean-evaluation drop. 

\emph{(iii) Inference-time unlearning.}
A third direction modifies the \emph{input} of inference time to make the model infer the input in an "unlearned" state.
Embedding-Corrupted (ECO) Prompts~\citep{liu2024large} train a token classifier to detect tokens to be unlearned, and then apply random noises in the embedding space of those tokens.
The random noise are optimized offline via zeroth-order procedures, and the resulting embedding-corrupted prompts are applied only at inference time, requiring no changes to LLM weights.
ECO has low inference cost but typically offers limited decontamination efficacy.
A key limitation is that the corruption is largely stochastic and provides limited directional control over memorized representations, thereby reducing effectiveness on strongly memorized items while still perturbing useful features on clean inputs.

\paragraph{Our approach}
Our method follows the white-box, inference-time paradigm, but intervenes exclusively at the input embedding level through a lightweight generator.
Compared with activation patching, we avoid directly editing internal circuits, reducing the risk of over-suppressing general capabilities.
Compared with coarse architectural bypassing, we provide a data-driven and instance-adaptive mechanism that targets contamination-induced behavior more selectively.
Compared with  ECO, DeconIEP learns instance-adaptive and directionally controlled perturbations,  avoiding purely stochastic corruption and eliminating the need for a token classifier.

Overall, this design yields competitive mitigation of contamination while substantially reducing performance sacrifice on clean benchmarks and incurring only small inference overhead.

\section{Limitations of ``Detect-then-Filter''}
\label{app:mia_limitation}
A widely considered mitigation approach is to first apply a membership
inference algorithm (MIA) to the benchmark and then evaluate only on
examples predicted as ``non-member.'' Let 
$D_{\text{test}} = D_{\text{clean}} \cup D_{\text{con}}$ with 
$D_{\text{clean}} \cap D_{\text{con}} = \emptyset$.
We model an arbitrary MIA detector as a binary classifier 
$\text{MIA}: \mathcal{X} \to \{0,1\}$:
\begin{numcases}{
\small
    \text{MIA}(x\in D_{\text{test}})=}
    0, & $x\in D_{\text{clean}}$,  \\
    1, & $x\in D_{\text{con}}$ .
\end{numcases}For this detector, we define
\begin{equation*}
\small
    \begin{aligned}
        &\text{FNR}
        =
        \Pr[\mathrm{MIA}(x)=\text{0} \mid x\in D_{\text{con}}],\\
        &\text{FPR}
        =
        \Pr[\mathrm{MIA}(x)=\text{1} \mid x\in D_{\text{clean}}],
    \end{aligned}
\end{equation*}
and retain after detection
\[
\small
D_{\text{neg}} = \{x \in D_{\text{test}} : \mathrm{MIA}(x)=\text{0}\},
\]
on which evaluation is performed.  
We denote the contaminated model by $f_{\text{con}}$, the clean model by
$f_{\text{clean}}$, and the contamination rate by
$\alpha = |D_{\text{con}}|/|D_{\text{test}}|$.

A simple counting argument yields the expected performance of
$f_{\text{con}}$ on $D_{\text{neg}}$:
\begin{equation}
\small
\begin{aligned}
      &\text{Perf}(D_{\text{neg}}, f_{\text{con}})=\\
  &\frac{
    (1-\alpha)(1-\text{FPR}) \,\text{Perf}(D_{\text{clean}}, f_{\text{con}})
    +
    \alpha \,\text{FNR} \,\text{Perf}(D_{\text{con}}, f_{\text{con}})
  }{
    (1-\alpha)(1-\text{FPR}) + \alpha \,\text{FNR}
  }
  \label{eq:mia_perf_neg_short}
\end{aligned}
\end{equation}

\paragraph{Derivation of Eq.~\eqref{eq:mia_perf_neg_short}}
We view performance as the average expected utility under the model output
distribution.

\begin{equation}
\small
\begin{aligned}
\mathrm{Perf}(S,f_{\mathrm{con}})
=
\frac{1}{|S|}
\sum_{x\in S}
\mathbb{E}_{y\sim p_{\mathrm{con}}(\cdot\mid x)}
\Bigl[u(x,y)\Bigr].
\end{aligned}
\label{eq:perf_def_u}
\end{equation}

We define the filtered (negative) subset retained by the detector.

\begin{equation}
\small
\begin{aligned}
D_{\text{neg}}
=
\Bigl\{
x\in D_{\text{test}}
:
\mathrm{MIA}(x)=0
\Bigr\}.
\end{aligned}
\label{eq:def_dneg}
\end{equation}

We parameterize the test set composition by the contamination rate and the
dataset size.

\begin{equation}
\small
\begin{aligned}
N
&=
|D_{\text{test}}|,\\
\alpha
&=
\frac{|D_{\text{con}}|}{|D_{\text{test}}|},\\
|D_{\text{clean}}|
&=
(1-\alpha)N,\\
|D_{\text{con}}|
&=
\alpha N.
\end{aligned}
\label{eq:alpha_counts}
\end{equation}

We recall the detector error rates.

\begin{equation}
\small
\begin{aligned}
\mathrm{FNR}
&=
\Pr\!\bigl[\mathrm{MIA}(x)=0 \mid x\in D_{\text{con}}\bigr],\\
\mathrm{FPR}
&=
\Pr\!\bigl[\mathrm{MIA}(x)=1 \mid x\in D_{\text{clean}}\bigr].
\end{aligned}
\label{eq:fnr_fpr}
\end{equation}

From the definitions above, the keep (negative) probabilities are

\begin{equation}
\small
\begin{aligned}
\Pr\!\bigl[x\in D_{\text{neg}}\mid x\in D_{\text{clean}}\bigr]
&=
\Pr\!\bigl[\mathrm{MIA}(x)=0\mid x\in D_{\text{clean}}\bigr]\\
&=
1-\mathrm{FPR},\\
\Pr\!\bigl[x\in D_{\text{neg}}\mid x\in D_{\text{con}}\bigr]
&=
\Pr\!\bigl[\mathrm{MIA}(x)=0\mid x\in D_{\text{con}}\bigr]\\
&=
\mathrm{FNR}.
\end{aligned}
\label{eq:keep_probs}
\end{equation}

Therefore, the expected counts of retained clean / contaminated examples are

\begin{equation}
\small
\begin{aligned}
\mathbb{E}\bigl[|D_{\text{neg}}\cap D_{\text{clean}}|\bigr]
&=
(1-\alpha)N(1-\mathrm{FPR}),\\
\mathbb{E}\bigl[|D_{\text{neg}}\cap D_{\text{con}}|\bigr]
&=
\alpha N\,\mathrm{FNR}.
\end{aligned}
\label{eq:expected_counts_parts}
\end{equation}

Summing yields the expected size of the filtered set.

\begin{equation}
\small
\begin{aligned}
\mathbb{E}\bigl[|D_{\text{neg}}|\bigr]
&=
(1-\alpha)N(1-\mathrm{FPR})
+
\alpha N\,\mathrm{FNR}.
\end{aligned}
\label{eq:expected_count_total}
\end{equation}

Define the per-input expected utility under the contaminated model.

\begin{equation}
\small
\begin{aligned}
\bar{u}_{\mathrm{con}}(x)
=
\mathbb{E}_{y\sim p_{\mathrm{con}}(\cdot\mid x)}
\Bigl[u(x,y)\Bigr].
\end{aligned}
\label{eq:u_bar_def}
\end{equation}

The total retained utility decomposes into two disjoint parts.

\begin{equation}
\small
\begin{aligned}
\sum_{x\in D_{\text{neg}}}\bar{u}_{\mathrm{con}}(x)
=
\sum_{x\in D_{\text{neg}}\cap D_{\text{clean}}}\bar{u}_{\mathrm{con}}(x)
+
\sum_{x\in D_{\text{neg}}\cap D_{\text{con}}}\bar{u}_{\mathrm{con}}(x).
\end{aligned}
\label{eq:sum_decompose}
\end{equation}

Assuming the detector induces \emph{composition-only} selection within each
split (i.e., retained examples are representative within $D_{\text{clean}}$
and within $D_{\text{con}}$), we have

\begin{equation}
\small
\begin{aligned}
\mathbb{E}\!\left[
\sum_{x\in D_{\text{neg}}\cap D_{\text{clean}}}\bar{u}_{\mathrm{con}}(x)
\right]
&=
\mathbb{E}\bigl[|D_{\text{neg}}\cap D_{\text{clean}}|\bigr]\,
\mathrm{Perf}(D_{\text{clean}},f_{\mathrm{con}}),\\
\mathbb{E}\!\left[
\sum_{x\in D_{\text{neg}}\cap D_{\text{con}}}\bar{u}_{\mathrm{con}}(x)
\right]
&=
\mathbb{E}\bigl[|D_{\text{neg}}\cap D_{\text{con}}|\bigr]\,
\mathrm{Perf}(D_{\text{con}},f_{\mathrm{con}}).
\end{aligned}
\label{eq:expected_sum_two_parts}
\end{equation}

Substituting Eq.~\eqref{eq:expected_counts_parts} gives the expected total
retained utility.

\begin{equation}
\small
\begin{aligned}
\mathbb{E}\!\left[
\sum_{x\in D_{\text{neg}}}\bar{u}_{\mathrm{con}}(x)
\right]
&=
(1-\alpha)N(1-\mathrm{FPR})\,
\mathrm{Perf}(D_{\text{clean}},f_{\mathrm{con}})\\
&\quad+
\alpha N\,\mathrm{FNR}\,
\mathrm{Perf}(D_{\text{con}},f_{\mathrm{con}}).
\end{aligned}
\label{eq:expected_total_util}
\end{equation}

By definition, the filtered performance is

\begin{equation}
\small
\begin{aligned}
\mathrm{Perf}(D_{\text{neg}},f_{\mathrm{con}})
=
\frac{1}{|D_{\text{neg}}|}
\sum_{x\in D_{\text{neg}}}\bar{u}_{\mathrm{con}}(x).
\end{aligned}
\label{eq:perf_dneg_def}
\end{equation}

Using a large-$N$ approximation (ratio of expectations), we obtain

\begin{equation}
\small
\begin{aligned}
\mathbb{E}\bigl[\mathrm{Perf}(D_{\text{neg}},f_{\mathrm{con}})\bigr]
&\approx
\frac{
\mathbb{E}\!\left[
\sum_{x\in D_{\text{neg}}}\bar{u}_{\mathrm{con}}(x)
\right]
}{
\mathbb{E}\bigl[|D_{\text{neg}}|\bigr]
}.
\end{aligned}
\label{eq:ratio_approx}
\end{equation}

Plugging Eq.~\eqref{eq:expected_total_util} and Eq.~\eqref{eq:expected_count_total},
and canceling the common factor $N$, yields

\begin{equation}
\small
\begin{aligned}
&\mathbb{E}\bigl[\mathrm{Perf}(D_{\text{neg}},f_{\mathrm{con}})\bigr]
\approx\\
&\frac{
(1-\alpha)(1-\mathrm{FPR})\,\mathrm{Perf}(D_{\text{clean}},f_{\mathrm{con}})
+
\alpha\,\mathrm{FNR}\,\mathrm{Perf}(D_{\text{con}},f_{\mathrm{con}})
}{
(1-\alpha)(1-\mathrm{FPR})
+
\alpha\,\mathrm{FNR}
},
\end{aligned}
\label{eq:mia_perf_neg_detailed}
\end{equation}
which matches Eq.~\eqref{eq:mia_perf_neg_short}.

\begin{figure}[htp]
    \centering
    \begin{subfigure}[t]{0.49\linewidth}
        \centering
        \includegraphics[width=\linewidth]{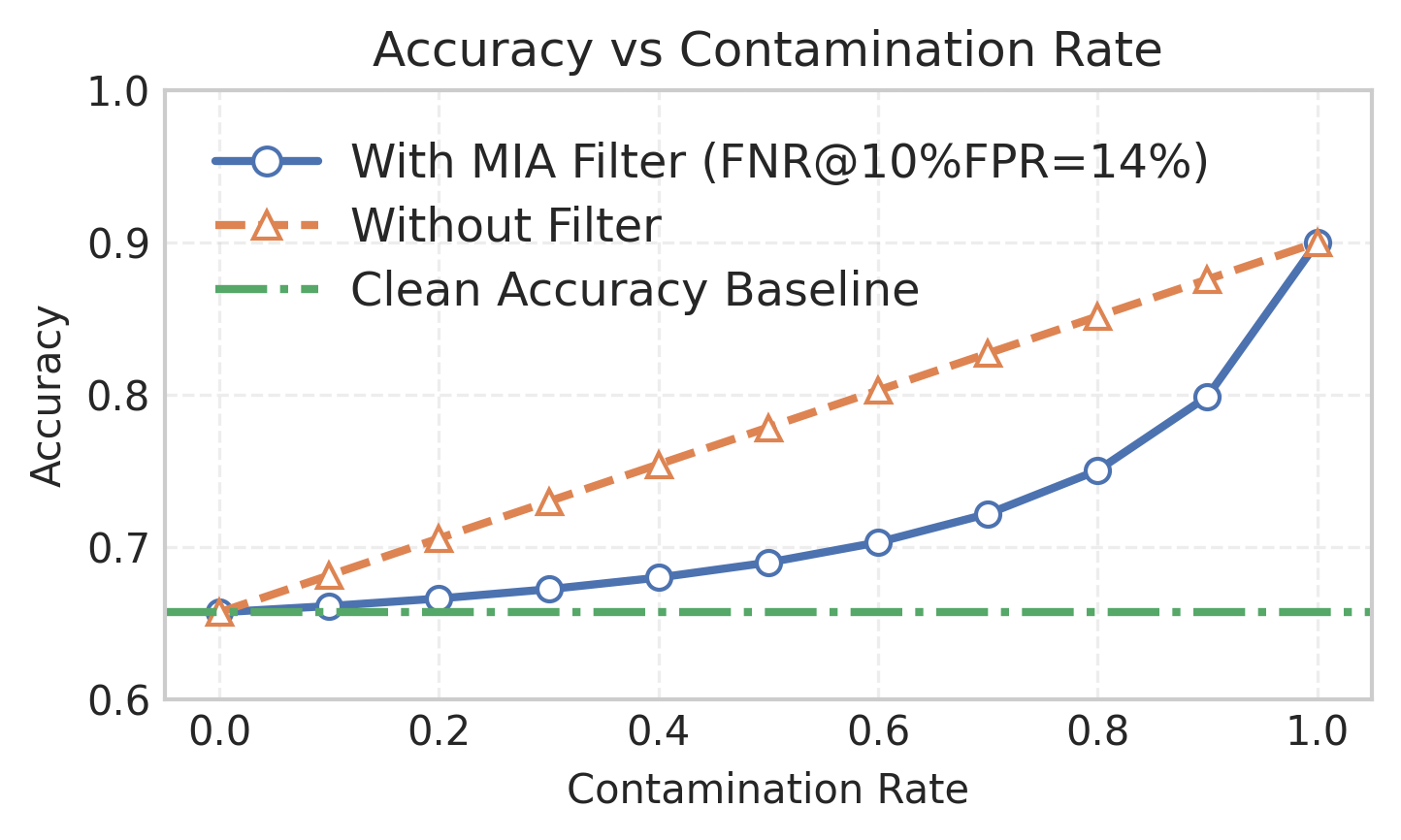}
        \caption{\textbf{Theoretical simulation.}}
        \label{fig:mia_mmlu_t}
    \end{subfigure}\hfill
    \begin{subfigure}[t]{0.49\linewidth}
        \centering
        \includegraphics[width=\linewidth]{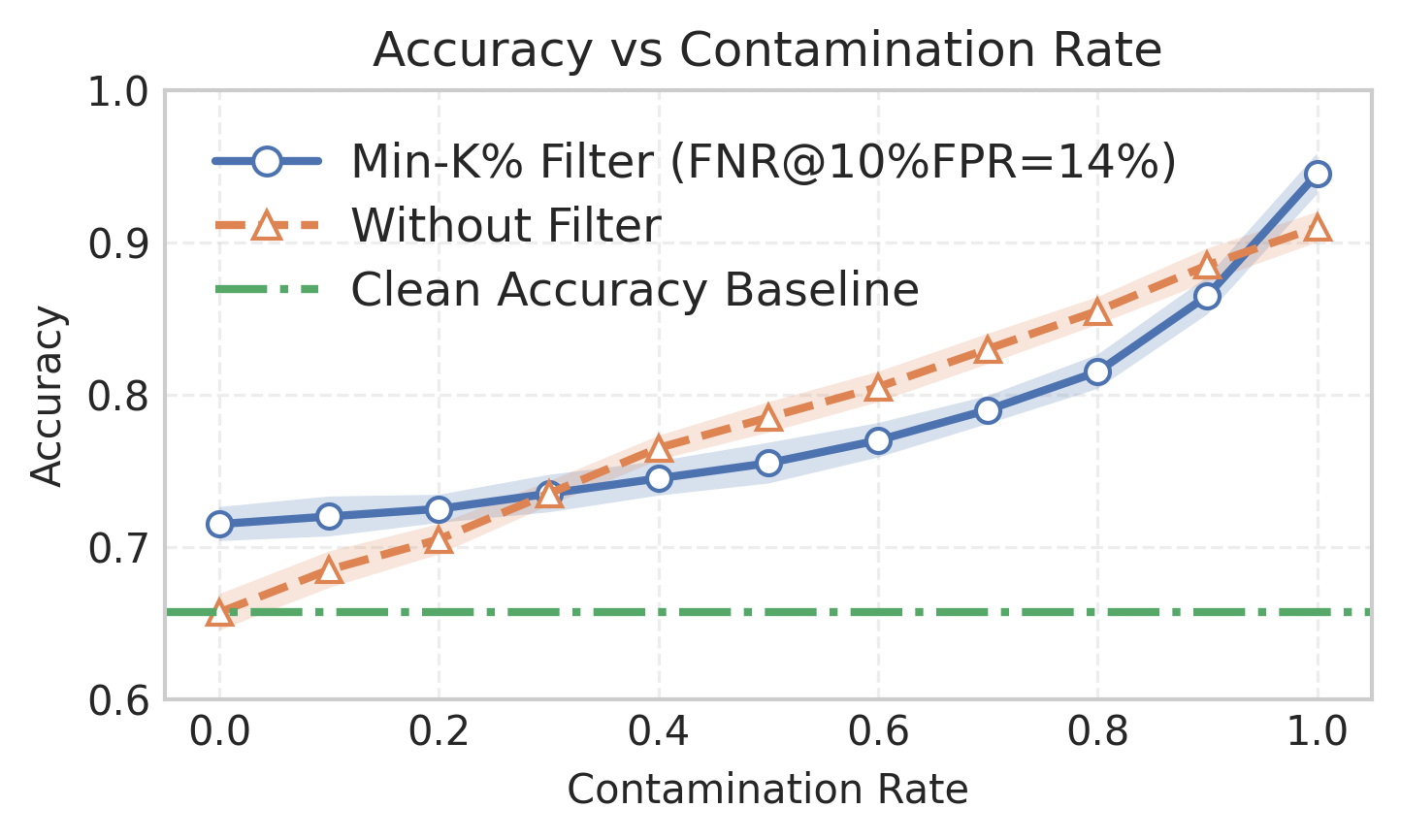}
        \caption{\textbf{Empirical measurement.}}
        \label{fig:mia_mmlu_emp}
    \end{subfigure}
    \caption{\textbf{Left:} theoretical simulation of Eq.~\ref{eq:mia_perf_neg_short} at a fixed detector operating point. \textbf{Right:} empirical MMLU results showing the same trend when applying a Min-K\% filter versus no filtering.}

    \label{fig:mia_mmlu}
\end{figure}

Equation~\ref{eq:mia_perf_neg_short} shows that \textbf{when contamination is non-trivial, detection-based filtering cannot
eliminate inflation: any non-zero FNR leaves enough contaminated
samples for inflation to persist as $\alpha$ increases.} Our simulation (Figure~\ref{fig:mia_mmlu_t}) and empirical results on an
artificially contaminated MMLU model (Figure~\ref{fig:mia_mmlu_emp}) confirm
this effect: even Min-K\% with $\mathrm{FNR}@10\%\mathrm{FPR}\approx 14\%$ in our
setting leaves substantial inflation as $\alpha$ increases, while other MIA
baselines (e.g., loss/perplexity thresholding and likelihood-ratio
methods~\citep{carlini2022membership,shi2023detecting}) typically exhibit higher
$\mathrm{FNR}@10\%\mathrm{FPR}$ in LLM settings.

\paragraph{Why strong MIA is still insufficient in practice}
Our results with Min-K\% indicate that even a strong LLM-oriented detector can
leave substantial residual inflation under moderate-to-high contamination.
This sensitivity is inherent to filtering: preventing inflation requires
extremely small $\text{FNR}$ at low $\text{FPR}$, which is difficult to achieve
robustly across models, benchmarks, and prompting formats.
Simpler detectors such as loss/perplexity thresholding and likelihood-ratio
variants often yield higher $\text{FNR}$ at comparable $\text{FPR}$ in LLM
settings, leading to even larger residual bias.

\paragraph{Selection bias and evaluation mismatch}
Detect-then-filter changes the evaluation distribution from $D_{\text{test}}$
to $D_{\text{neg}}$. Even if the goal is to approximate clean evaluation,
discarding items can bias the reported score because the retained subset is not
guaranteed to be representative of the original benchmark, especially when
$\text{FPR}$ is non-negligible.

\paragraph{Implications for benchmark contamination}
While extreme leakage may be rare at the full-corpus level, widely reused
benchmarks can accumulate exposure through repeated fine-tuning, dataset mixing,
and web-crawled mirrors. Consequently, even moderate overall leakage can induce
non-trivial effective contamination for popular subsets, where false negatives
can dominate the residual inflation. This motivates inference-time
interventions beyond filtering.

\begin{figure*}[t]
    \centering
    \includegraphics[width=0.8\linewidth]{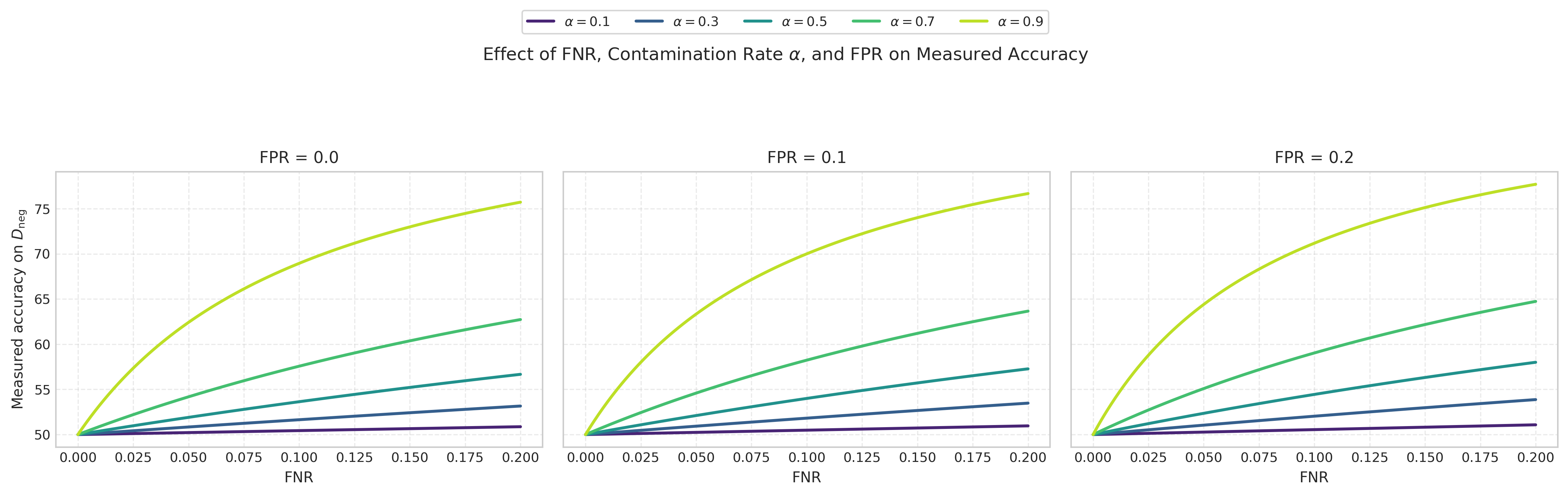}
    \caption{\textbf{Simulated effect of MIA errors.}
    Measured accuracy of the contaminated model $f_{\text{con}}$ on
    $D_{\text{neg}}$ as a function of FNR, for different contamination
    rates $\alpha$ and FPR values, computed from
    Equation~\eqref{eq:mia_perf_neg_short} with
    $\text{Perf}(D_{\text{clean}}, f_{\text{con}})=50\%$ and
    $\text{Perf}(D_{\text{con}}, f_{\text{con}})=90\%$.  
    Even small FNR values lead to substantial inflation when $\alpha$ is
    large.}
    \label{fig:mia_sim}
\end{figure*}

\paragraph{More simulation of MIA }
Figure~\ref{fig:mia_sim} visualizes Equation~\eqref{eq:mia_perf_neg_short}
for different contamination rates $\alpha$ and false positive rates FPR.
Even with a modest FNR on the x-axis, the measured accuracy on
$D_{\text{neg}}$ quickly departs from the clean baseline once $\alpha$ is
moderate or large.  

\section{Theorem \ref{thm:kl-bound}}
\label{app:kl-bound}
\begin{theorem}[KL upper bound on performance gap]
\label{thm:kl-bound}
Fix a test set $D_{\mathrm{test}}$ and suppose
\[
\mathrm{Perf}(f, D_{\mathrm{test}})
\;=\;
\mathbb{E}_{x\in D_{\mathrm{test}}}\;
\mathbb{E}_{y\sim p_f(\cdot\mid x)}[u(x,y)]
\]
for some utility $u(x,y)\in[0,1]$.
Then, $\Delta_{\mathrm{perf}}$ is bounded by:
\begin{equation}
\Delta_{\mathrm{perf}}
\le
\sqrt{\frac{1}{2}
\mathbb{E}_{x}
\Big[
\mathrm{KL}\big(
p_{\mathrm{con}}(\cdot|x)\,\|\,p_{\mathrm{clean}}(\cdot|x)
\big)
\Big]}.
\end{equation}
In particular, if the average KL divergence between
$p_{\mathrm{con}}(\cdot\mid x)$ and $p_{\mathrm{clean}}(\cdot\mid x)$ is at
most $\varepsilon$, then
$\Delta_{\mathrm{perf}} \le \sqrt{\varepsilon/2}$.
\end{theorem}

\label{app:proof}
\begin{proof}[Proof ]
For fixed $x$ and any bounded $u(x,\cdot)\in[0,1]$, standard arguments imply
that the difference in expected utility is bounded by the total variation (TV)
distance:
\begin{align}
\bigl|
\mathbb{E}_{p_{\mathrm{con}}}&[u(x,\cdot)]
-
\mathbb{E}_{p_{\mathrm{clean}}}[u(x,\cdot)]
\bigr| \nonumber \\
&\le
\mathrm{TV}\big(p_{\mathrm{con}}(\cdot\mid x), p_{\mathrm{clean}}(\cdot\mid x)\big).
\end{align}
Averaging over $x\in D_{\mathrm{test}}$ yields an upper bound on
$\Delta_{\mathrm{perf}}$.
Applying Pinsker's inequality,
$\mathrm{TV}(P,Q) \le \sqrt{\tfrac12\,\mathrm{KL}(P\|Q)}$, and then Jensen's
inequality for the concave square-root function gives the stated bound in
terms of the average KL divergence.
\end{proof}

\section{Generator Architecture Details}
\label{app:generator-arch}

\paragraph{Overview}
Our perturbation generator consists of a network ${G}_{\theta}$ followed by a deterministic bounding layer.
Given an input prompt $x$ and its embedding sequence $e(x)\in\mathbb{R}^{L\times d}$ (computed by the contaminated model tokenizer/embedding layer), the generator first produces a raw perturbation
\begin{equation}
\tilde{\delta}(x)={G}_{\theta}(e(x))\in\mathbb{R}^{L\times d},
\end{equation}
and then applies a scaled $\tanh$ to obtain the bounded perturbation
\begin{equation}
\delta(x)=\zeta\cdot\tanh\!\bigl(\tilde{\delta}(x)\bigr),
\end{equation}
which enforces $\|\delta(x)\|_{\infty}\le \zeta$ by construction. The perturbed embedding
$e(x)+\delta(x)$ is then fed to $f_{\mathrm{con}}$ at inference time.

\paragraph{Unconstrained generator ${G}_{\theta}$}
We instantiate ${G}_{\theta}$ as a lightweight decoder-only Transformer with $n$ layers.
Let $z=e(x)\in\mathbb{R}^{L\times d}$ be the input embedding sequence.
We apply an input projection, $n$ residual decoder blocks, and an output head:
\begin{equation}
\begin{aligned}
H_{0} &= z W_{\mathrm{in}}, \\
H_{\ell} &= \mathrm{Block}_{\ell}(H_{\ell-1}) \quad (\ell=1,\dots,n), \\
\tilde{\delta}(x) &= H_{n} W_{\mathrm{out}},
\end{aligned}
\end{equation}
where $W_{\mathrm{in}}\in\mathbb{R}^{d\times d_g}$ and $W_{\mathrm{out}}\in\mathbb{R}^{d_g\times d}$ are learnable matrices and $d_g$ is the hidden width of the generator (we use $d_g=d$ by default). The output $\tilde{\delta}(x)$ has the same shape as $e(x)$ to support token-wise perturbations.

\paragraph{Decoder block}
Each decoder block uses standard pre-norm Transformer components:
\begin{equation}
\begin{aligned}
U_{\ell} &= H_{\ell-1} + \mathrm{MHA}\bigl(\mathrm{LN}(H_{\ell-1})\bigr),\\
H_{\ell} &= U_{\ell} + \mathrm{FFN}\bigl(\mathrm{LN}(U_{\ell})\bigr),
\end{aligned}
\end{equation}
where $\mathrm{MHA}$ is masked multi-head self-attention, $\mathrm{FFN}$ is a position-wise feed-forward network, and $\mathrm{LN}$ is LayerNorm. We use the same causal attention mask as the base LLM so that the generator is compatible with autoregressive prompting.

\paragraph{Implementation choices}
Unless otherwise specified, we set $n=4$ and use dropout $0.2$ within the generator blocks. We found that this lightweight instantiation is sufficient to amortize per-instance perturbation optimization; our main results are not intended to depend on a sophisticated generator architecture, but rather on the reference-guided objective and the bounded embedding-space intervention.

\paragraph{Discussion}
The $\tanh$ bounding ensures a strict $\ell_{\infty}$ constraint and keeps the intervention local in embedding space. Alternative choices such as hard clipping yield similar behavior but introduce non-smooth gradients during training; we adopt the scaled $\tanh$ for stable optimization.

\begin{algorithm}[h]
\small
\caption{Training of the Noise Generator $G_{\theta}$}
\label{alg:train_gen}
\KwIn{%
  Auxiliary dataset $D_{\text{aux}}$; contaminated model $f_{\text{con}}$; reference model $f_{\text{ref}}$; \\
  learning rate $\eta>0$; max norm $\zeta>0$; \\
  Max iterations $T\in\mathbb{N}$; batch size $B\in\mathbb{N}$; optional stop threshold $\epsilon>0$
}
\KwOut{Trained parameters $\theta$ of $G_{\theta}$}

\BlankLine
Initialize $\theta$\;

\For{$t = 1$ \KwTo $T$}{
  Sample minibatch $\mathcal{B}=\{x_i\}_{i=1}^{B} \subset D_{\text{aux}}$\;

  \For{$i=1$ \KwTo $B$}{
    $z_i \leftarrow e(x_i)$ \tcp*{token embeddings}
    $\delta_i \leftarrow \zeta \cdot \tanh\!\big(G_{\theta}(z_i)\big)$ \tcp*{$\ell_{\infty}$-budgeted perturbation}
    $p^{\text{ref}}_i \leftarrow f_{\text{ref}}(z_i)$ \tcp*{reference output}
    $p^{\text{con}}_i \leftarrow f_{\text{con}}(z_i+\delta_i)$ \tcp*{perturbed contaminated output}
  }

  $\mathcal{L} \leftarrow \frac{1}{B}\sum_{i=1}^{B}\Big[
  \lambda_{\mathrm{KL}}\,\mathrm{KL}\!\big(p^{\text{con}}_i \,\|\, p^{\text{ref}}_i\big)
  +\lambda_{\mathrm{CE}}\,\mathrm{CE}\!\big(p^{\text{con}}_i,\, y^{\text{ref}}_i\big)
  \Big]$ \tcp*{$y^{\text{ref}}_i$ = hard labels from $f_{\text{ref}}$}

  $\theta \leftarrow \theta - \eta\,\nabla_{\theta}\mathcal{L}$ \tcp*{gradient descent}
  \If{ $\mathcal{L} \le \epsilon$}{\textbf{break}}
}

\Return{$\theta$}
\end{algorithm}

\begin{table*}[hbt]
\centering
\scriptsize
\setlength{\tabcolsep}{3.2pt}
\renewcommand{\arraystretch}{1.15}
\begin{tabular}{@{}llllcccccc@{}}
\toprule
\multirow{2}{*}{Model} & \multirow{2}{*}{Split} & \multirow{2}{*}{Method} & \multirow{2}{*}{Access} &
\multicolumn{3}{c}{TruthfulQA ($o=1$)} & \multicolumn{3}{c}{MMLU ($o=1$)} \\
\cmidrule(lr){5-7}\cmidrule(lr){8-10}
 &  &  &  & Exact & Semantic-level & Domain-level & Exact & Semantic-level & Domain-level \\
\midrule

\multirow{7}{*}{Mistral}
 & $f_{\mathrm{clean}}$ & baseline & \multicolumn{1}{c}{--} &
0.287 & 0.249 & 0.269 & 0.451 & 0.434 & 0.406 \\
\cmidrule(lr){2-10}
 & \multirow{6}{*}{$f_{\mathrm{con}}$} & W/O Decontamination & \multicolumn{1}{c}{--} &
0.890 (0.603) & 0.890 (0.641) & 0.865 (0.596) & 0.617 (0.166) & 0.601 (0.167) & 0.492 (0.086) \\
 &  & TED & Black-box &
0.785 (0.498) & 0.770 (0.521) & 0.736 (0.467) & \underline{0.404 (0.047)} & \textbf{0.405 (0.029)} & \textbf{0.363 (0.043)} \\
 &  & ITD & Black-box &
0.890 (0.603) & 0.890 (0.641) & 0.865 (0.596) & 0.610 (0.159) & 0.588 (0.154) & 0.499 (0.093) \\
 &  & Shortcut Neuron & White-box &
\underline{0.630 (0.343)} & \underline{0.639 (0.390)} & \underline{0.640 (0.371)} & 0.300 (0.151) & 0.301 (0.133) & 0.263 (0.143) \\
 &  & Short Circuit & White-box &
0.890 (0.603) & 0.880 (0.631) & 0.827 (0.558) & 0.631 (0.180) & 0.623 (0.189) & 0.594 (0.188) \\
 &  & DeconIEP(Ours) & White-box &
\textbf{0.366 (0.079)} & \textbf{0.322 (0.073)} & \textbf{0.408 (0.139)} & \textbf{0.478 (0.027)} & \underline{0.467 (0.033)} & \underline{0.470 (0.064)} \\
\midrule

\multirow{7}{*}{Qwen 2.5}
 & $f_{\mathrm{clean}}$ & baseline & \multicolumn{1}{c}{--} &
0.555 & 0.531 & 0.572 & 0.670 & 0.671 & 0.673 \\
\cmidrule(lr){2-10}
 & \multirow{6}{*}{$f_{\mathrm{con}}$} & W/O Decontamination & \multicolumn{1}{c}{--} &
0.866 (0.311) & 0.856 (0.325) & 0.784 (0.212) & 0.775 (0.105) & 0.769 (0.098) & 0.735 (0.062) \\
 &  & TED & Black-box &
0.789 (0.234) & 0.813 (0.282) & 0.760 (0.188) & 0.742 (0.072) & \underline{0.632 (0.039)} & \textbf{0.619 (0.054)} \\
 &  & ITD & Black-box &
0.856 (0.301) & 0.856 (0.325) & 0.774 (0.202) & 0.776 (0.106) & 0.766 (0.095) & 0.742 (0.069) \\
 &  & Shortcut Neuron & White-box &
0.306 (0.249) & 0.297 (0.234) & 0.274 (0.298) & 0.395 (0.275) & 0.417 (0.254) & 0.389 (0.284) \\
 &  & Short Circuit & White-box &
\textbf{0.602 (0.047)} & \textbf{0.659 (0.128)} & \textbf{0.609 (0.037)} & \textbf{0.619 (0.051)} & \textbf{0.620 (0.051)} & 0.610 (0.063) \\
 &  & DeconIEP(Ours) & White-box &
\underline{0.618 (0.063)} & \underline{0.664 (0.133)} & \underline{0.655 (0.083)} & \underline{0.705 (0.035)} & 0.725 (0.054) & \underline{0.730 (0.057)} \\
\midrule

\multirow{7}{*}{LLaMA-3}
 & $f_{\mathrm{clean}}$ & baseline & \multicolumn{1}{c}{--} &
0.474 & 0.467 & 0.452 & 0.576 & 0.565 & 0.549 \\
\cmidrule(lr){2-10}
 & \multirow{6}{*}{$f_{\mathrm{con}}$} & W/O Decontamination & \multicolumn{1}{c}{--} &
0.900 (0.426) & 0.909 (0.442) & 0.885 (0.433) & 0.734 (0.159) & 0.728 (0.163) & 0.663 (0.114) \\
 &  & TED & Black-box &
0.886 (0.412) & 0.876 (0.409) & 0.851 (0.399) & 0.727 (0.151) & 0.735 (0.170) & \textbf{0.576 (0.027)} \\
 &  & ITD & Black-box &
0.876 (0.402) & 0.876 (0.409) & 0.856 (0.404) & 0.721 (0.145) & 0.683 (0.118) & 0.620 (0.070) \\
 &  & Shortcut Neuron & White-box &
0.804 (0.330) & 0.789 (0.323) & 0.644 (0.192) & \textbf{0.547 (0.029)} & \textbf{0.566 (0.001)} & 0.484 (0.065) \\
 &  & Short Circuit & White-box &
\underline{0.589 (0.115)} & \underline{0.612 (0.146)} & \underline{0.635 (0.183)} & 0.695 (0.120) & 0.682 (0.117) & 0.710 (0.160) \\
 &  & DeconIEP(Ours) & White-box &
\textbf{0.517 (0.043)} & \textbf{0.476 (0.009)} & \textbf{0.461 (0.009)} &
\underline{0.616 (0.040)} & \underline{0.525 (0.041)} & \underline{0.515 (0.034)} \\
\bottomrule
\end{tabular}
\caption{Decontamination results for $o=1$ across different models. Values denote accuracy, with parentheses showing Residual Contamination (RC) (smaller is better). Best RC is in bold and second best is underlined within each model and split.}
\label{tab:decontam_o1_all}
\end{table*}
\begin{table*}[hbt]
\centering
\scriptsize
\setlength{\tabcolsep}{3.2pt}
\renewcommand{\arraystretch}{1.15}
\begin{tabular}{@{}llllcccccc@{}}
\toprule
\multirow{2}{*}{Model} & \multirow{2}{*}{Split} & \multirow{2}{*}{Method} & \multirow{2}{*}{Access} &
\multicolumn{3}{c}{TruthfulQA ($o=5$)} & \multicolumn{3}{c}{MMLU ($o=5$)} \\
\cmidrule(lr){5-7}\cmidrule(lr){8-10}
 &  &  &  & Exact & Semantic-level & Domain-level & Exact & Semantic-level & Domain-level \\
\midrule

\multirow{7}{*}{Mistral}
 & $f_{\mathrm{clean}}$ & baseline & \multicolumn{1}{c}{--} &
0.287 & 0.249 & 0.269 & 0.451 & 0.434 & 0.406 \\
\cmidrule(lr){2-10}
 & \multirow{6}{*}{$f_{\mathrm{con}}$} & W/O Decontamination & \multicolumn{1}{c}{--} &
0.995 (0.708) & 1.000 (0.751) & 0.885 (0.616) & 0.995 (0.544) & 0.972 (0.538) & 0.463 (0.057) \\
 &  & TED & Black-box &
\textbf{0.086 (0.201)} & \textbf{0.244 (0.005)} & 0.712 (0.443) & 0.074 (0.377) & 0.324 (0.110) & \underline{0.383 (0.023)} \\
 &  & ITD & Black-box &
0.995 (0.708) & 1.000 (0.751) & 0.885 (0.616) & 0.994 (0.543) & 0.967 (0.533) & 0.467 (0.061) \\
 &  & Shortcut Neuron & White-box &
0.621 (0.334) & 0.601 (0.352) & \underline{0.640 (0.371)} & \underline{0.394 (0.057)} & \underline{0.409 (0.025)} & 0.245 (0.161) \\
 &  & Short Circuit & White-box &
0.919 (0.632) & 0.914 (0.665) & 0.736 (0.467) & 0.581 (0.130) & 0.535 (0.101) & \textbf{0.407 (0.001)} \\
 &  & DeconIEP(Ours) & White-box &
\underline{0.595 (0.308)} & \underline{0.551 (0.302)} & \textbf{0.570 (0.301)} &
\textbf{0.454 (0.003)} & \textbf{0.423 (0.011)} & 0.433 (0.027) \\
\midrule

\multirow{7}{*}{Qwen 2.5}
 & $f_{\mathrm{clean}}$ & baseline & \multicolumn{1}{c}{--} &
0.555 & 0.531 & 0.572 & 0.670 & 0.671 & 0.673 \\
\cmidrule(lr){2-10}
 & \multirow{6}{*}{$f_{\mathrm{con}}$} & W/O Decontamination & \multicolumn{1}{c}{--} &
1.000 (0.445) & 0.981 (0.450) & 0.928 (0.356) & 0.981 (0.311) & 0.942 (0.271) & 0.740 (0.067) \\
 &  & TED & Black-box &
0.191 (0.364) & \underline{0.632 (0.101)} & 0.788 (0.216) & 0.870 (0.200) & 0.837 (0.166) & 0.886 (0.213) \\
 &  & ITD & Black-box &
1.000 (0.445) & 0.981 (0.450) & 0.928 (0.356) & 0.983 (0.313) & 0.938 (0.267) & 0.752 (0.079) \\
 &  & Shortcut Neuron & White-box &
\textbf{0.517 (0.038)} & \textbf{0.507 (0.024)} & \underline{0.442 (0.130)} & 0.405 (0.265) & 0.419 (0.252) & 0.359 (0.314) \\
 &  & Short Circuit & White-box &
\underline{0.650 (0.095)} & 0.636 (0.105) & \textbf{0.677 (0.105)} & \textbf{0.750 (0.080)} & \underline{0.753 (0.082)} & \underline{0.707 (0.034)} \\
 &  & DeconIEP(Ours) & White-box &
0.790 (0.235) & 0.799 (0.268) & 0.718 (0.146) & \underline{0.805 (0.135)} & \textbf{0.725 (0.054)} & \textbf{0.641 (0.032)} \\
\midrule

\multirow{7}{*}{LLaMA-3}
 & $f_{\mathrm{clean}}$ & baseline & \multicolumn{1}{c}{--} &
0.474 & 0.467 & 0.452 & 0.576 & 0.565 & 0.549 \\
\cmidrule(lr){2-10}
 & \multirow{6}{*}{$f_{\mathrm{con}}$} & W/O Decontamination & \multicolumn{1}{c}{--} &
0.995 (0.522) & 0.990 (0.524) & 0.923 (0.471) & 0.992 (0.416) & 0.961 (0.395) & 0.641 (0.092) \\
 &  & TED & Black-box &
0.866 (0.392) & 0.866 (0.399) & 0.870 (0.418) & 0.794 (0.218) & 0.785 (0.220) & \underline{0.610 (0.061)} \\
 &  & ITD & Black-box &
0.909 (0.435) & 0.928 (0.462) & 0.870 (0.418) & 0.824 (0.248) & 0.828 (0.263) & \underline{0.610 (0.061)} \\
 &  & Shortcut Neuron & White-box &
0.852 (0.378) & 0.856 (0.390) & 0.726 (0.274) & \underline{0.705 (0.130)} & \underline{0.685 (0.120)} & 0.427 (0.122) \\
 &  & Short Circuit & White-box &
\textbf{0.589 (0.115)} & \underline{0.622 (0.155)} & \underline{0.606 (0.154)} &
\textbf{0.697 (0.121)} & 0.695 (0.130) & 0.706 (0.157) \\
 &  & DeconIEP(Ours) & White-box &
\underline{0.660 (0.187)} & \textbf{0.562 (0.096)} & \textbf{0.572 (0.120)} &
0.736 (0.160) & \textbf{0.527 (0.038)} & \textbf{0.527 (0.022)} \\
\bottomrule
\end{tabular}
\caption{Decontamination results for $o=5$ across different models. Values denote accuracy, with parentheses showing Residual Contamination (RC) (smaller is better). Best RC is in bold and second best is underlined within each model and split.}
\label{tab:decontam_o5_all}
\end{table*}

\section{More Experiment}
\label{app:more experimrnt}
\subsection{Definition of RC and BUD}
\label{app: Definition of RC and BUD}
We quantify the \emph{decontamination effect} and the \emph{benign side effect} of
an inference-time mitigation operator $M(\cdot)$ using two complementary metrics.
Let $f_{\mathrm{con}}$ denote the contaminated model under evaluation, and
$f_{\mathrm{clean}}$ denote a reference model of the
same architecture. Let $D_{\mathrm{con}}$ be a contaminated (seen or near-seen)
evaluation set, and $D_{\mathrm{ clean}}$ be an uncontaminated evaluation set.

Formally,
\[
\begin{aligned}
\mathrm{RC}(M)
&= \Bigl|
   \mathrm{Perf}(f_{\mathrm{clean}},D_{\mathrm{con}})
   \\
&\quad-
   \mathrm{Perf}(M(f_{\mathrm{con}}),D_{\mathrm{con}})
   \Bigr|, \\
\mathrm{BUD}(M)
&= \Bigl|
   \mathrm{Perf}(f_{\mathrm{con}},D_{\mathrm{uncon}})
   \\
&\quad-
   \mathrm{Perf}(M(f_{\mathrm{con}}),D_{\mathrm{uncon}})
   \Bigr|.
\end{aligned}
\]

\subsection{The Full Result on Different Models under Different $o$}
\label{app:The Full Result}

\paragraph{Evaluation splits (Exact / Semantic-level / Domain-level)}
For each benchmark, we report accuracy on three evaluation splits designed to
separate memorization from generalization.
\textbf{Exact} contains items that are exact matches (or maximally close
duplicates) of leaked training instances, thus directly reflecting
memorization-driven gains.
\textbf{Semantic-level} contains meaning-preserving rewrites of the leaked items
(e.g., paraphrases) that aim to break surface-form matching while retaining
the same answer and comparable difficulty, thereby probing semantic memorization
beyond verbatim recall.
\textbf{Domain-level} consists of in-domain but non-leaked instances sampled to
match the benchmark distribution, serving as an uncontaminated proxy for benign
generalization.

\paragraph{How to read Tables~\ref{tab:decontam_o1_all} and~\ref{tab:decontam_o5_all}}
Rows are grouped by backbone model. Within each model, the $f_{\mathrm{con}}$
block reports the contaminated model under different inference-time mitigation
methods, while the bottom row reports $f_{\mathrm{clean}}$ as a reference anchor.
Each entry shows accuracy; the value in parentheses is \textbf{Residual
Contamination (RC)}, computed as the absolute performance gap
$\bigl|\mathrm{Perf}(f_{\mathrm{con}})-\mathrm{Perf}(f_{\mathrm{clean}})\bigr|$
on the same split. Smaller RC indicates that the method more effectively removes
contamination-induced inflation (i.e., the mitigated model behaves closer to the
clean reference on contaminated splits).

\paragraph{Table~\ref{tab:decontam_o1_all} ($o=1$): mild leakage regime}
When contamination severity is mild, W/O and rewriting-based baselines often
retain elevated accuracy on \textbf{Exact} (and sometimes \textbf{Semantic-level}),
indicating that memorization signals remain largely intact.
DeconIEP consistently reduces \textbf{Exact} and \textbf{Semantic-level} accuracy
toward the $f_{\mathrm{clean}}$ level, while keeping \textbf{Domain-level}
accuracy comparatively stable. This pattern suggests a selective suppression of
memorization-driven behavior without broadly impairing in-domain generalization.
In contrast, activation patching and architectural bypassing can reduce
\textbf{Exact} as well, but their \textbf{Domain-level} degradation can be larger
or less predictable, consistent with heavier interventions that overlap with
general computation.

\paragraph{Table~\ref{tab:decontam_o5_all} ($o=5$): strong memorization regime}
Under severe leakage, W/O frequently becomes near-saturated on \textbf{Exact}
(and often remains high on \textbf{Semantic-level}), implying strong retrieval
of leaked solutions. In this regime, the key question is whether a method can
substantially reduce RC on contaminated splits without collapsing
\textbf{Domain-level} performance.
DeconIEP maintains a controlled reduction of contamination inflation (lower RC
on \textbf{Exact}/\textbf{Semantic-level}) while avoiding catastrophic drops on
\textbf{Domain-level}. Some baselines may become unstable or overly aggressive:
stochastic calibration (TED) can severely hurt accuracy on certain settings,
and broad architectural edits may trade stronger suppression for larger benign
utility loss. ``--'' denotes missing runs for the corresponding method/model.

\subsection{The result on Code Generation Task}

\label{app:codegen}

\begin{figure}[t]
    \centering
    \begin{subfigure}[t]{0.95\linewidth}
        \centering
        \includegraphics[width=\linewidth]{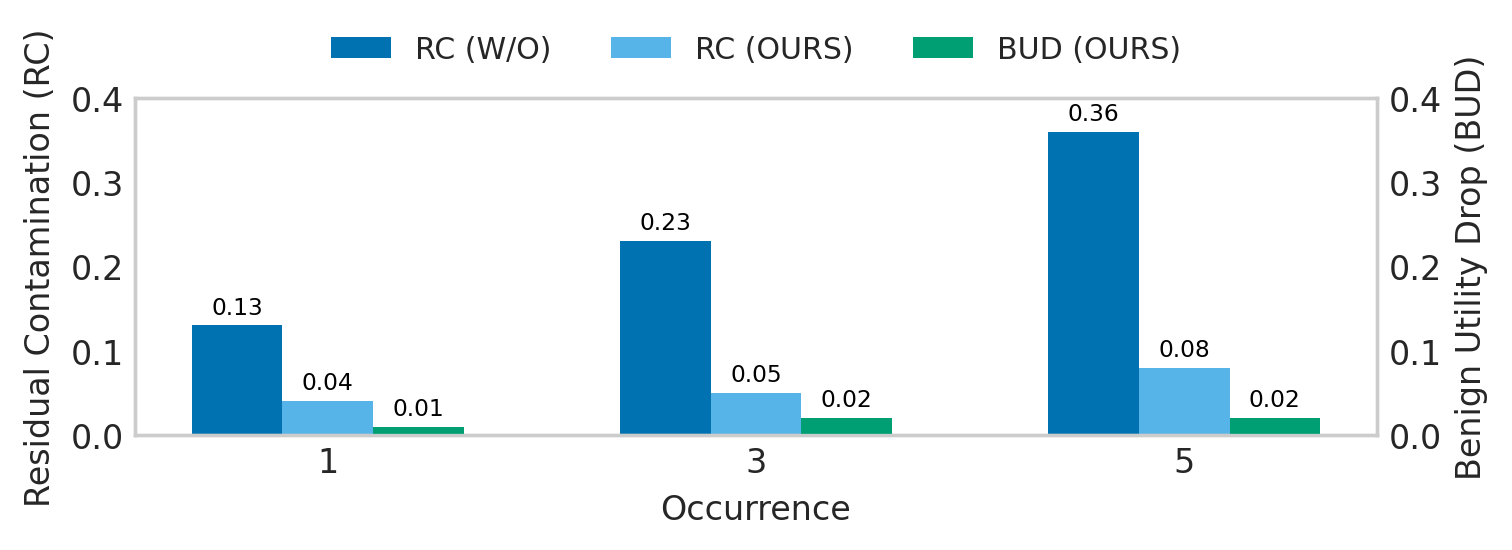}
        \caption{Ours}
        \label{fig:scale_truthfulqa}
    \end{subfigure}

    \vspace{0.35em}

    \begin{subfigure}[t]{0.95\linewidth}
        \centering
        \includegraphics[width=\linewidth]{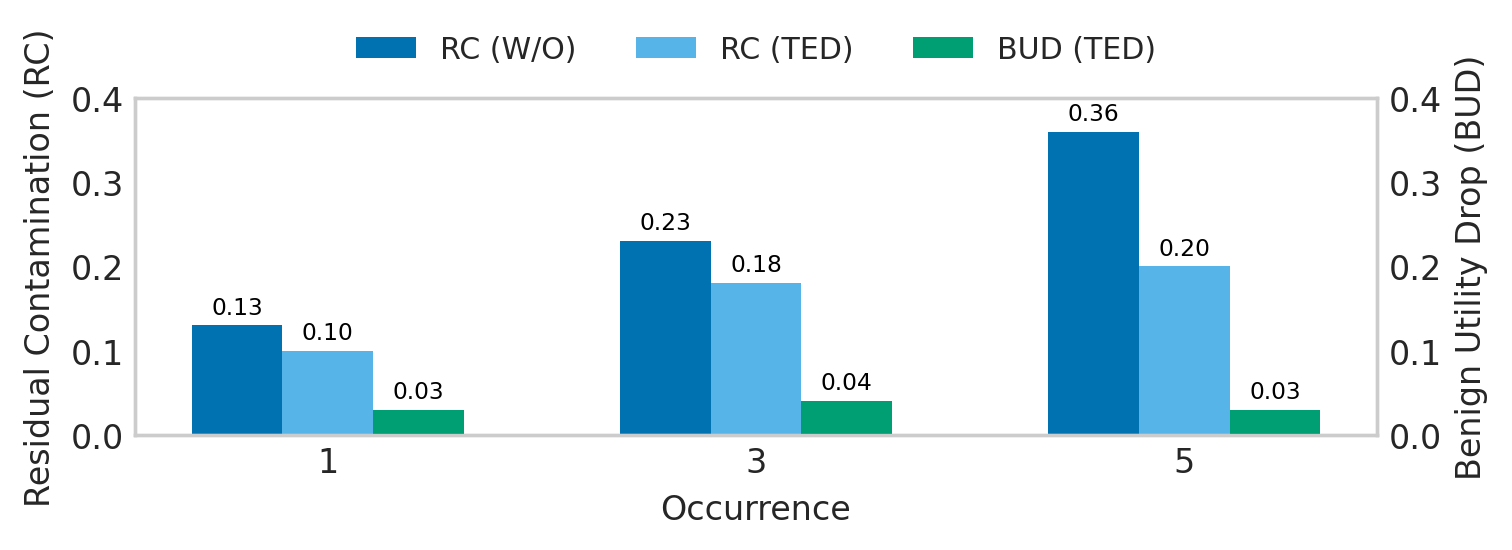}
        \caption{TED}
        \label{fig:scale_tqa}
    \end{subfigure}

\caption{\textbf{HumanEval (0-shot Pass@1) under increasing contamination occurrence $o$.}
We report residual contamination (RC; left axis, lower is better) and benign utility drop (BUD; right axis, lower is better) for {LLaMA-3-8b-instruct}. Compared to no mitigation, TED reduces RC with non-trivial BUD, while our method achieves larger RC reduction with smaller BUD across $o\in\{1,3,5\}$.}

    \label{fig:humaneval_occ}
\end{figure}

Due to limited computational resources, we evaluate code-generation performance
only on {LLaMA-3-8b-instruct} using HumanEval with 0-shot \textsc{Pass@1}.
Figure~\ref{fig:humaneval_occ} reports residual contamination (RC; lower is
better) and benign utility drop (BUD; lower is better) as the contamination
occurrence $o\in\{1,3,5\}$ increases.

Without mitigation, RC increases monotonically with $o$ (0.13$\rightarrow$0.23$\rightarrow$0.36), indicating stronger leakage effects under repeated exposure.
Both TED and our method reduce RC across all $o$, but with markedly different
trade-offs.
TED yields only moderate RC reduction (0.10, 0.18, 0.20) and incurs non-trivial
BUD (0.03--0.04).
In contrast, our method substantially suppresses RC (0.04, 0.05, 0.08), achieving
a larger absolute RC reduction (e.g., 0.36$\rightarrow$0.08 at $o=5$) while
maintaining consistently smaller BUD (0.01--0.02).
Overall, these results suggest that our inference-time perturbation better
removes contamination-driven shortcuts while preserving benign code-generation
utility under the same 0-shot Pass@1 evaluation.

\subsection{RC–BUD Pareto trade-off}

\label{app:pareto}

\begin{figure}[t]
  \centering
  \includegraphics[width=1.00\linewidth]{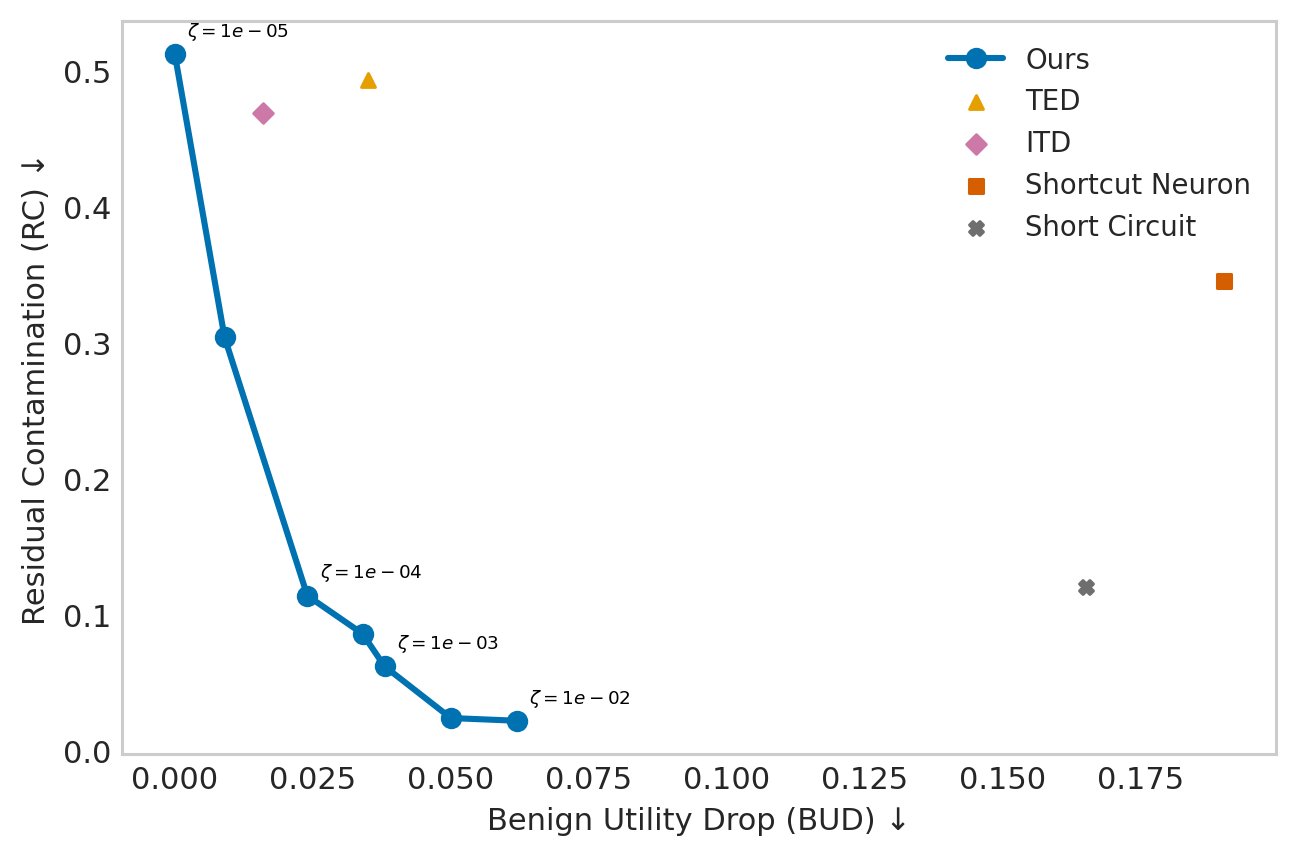}
  \caption{\textbf{RC--BUD Pareto trade-off induced by $\zeta$.}
  Each point on the DeconIEP curve corresponds to a different perturbation budget
  $\zeta$, illustrating controllable trade-offs between decontamination strength
  (RC) and benign utility drop (BUD). Baseline methods are shown as single points.}
  \label{fig:pareto-zeta}
\end{figure}

\paragraph{What Figure~\ref{fig:pareto-zeta} shows}
Figure~\ref{fig:pareto-zeta} visualizes the RC--BUD trade-off across methods.
Each baseline appears as a \emph{single point} because it does not expose a
continuous knob analogous to our perturbation budget. DeconIEP forms a
\emph{curve} because varying $\zeta$ continuously changes the intervention
strength.

\paragraph{Interpreting the axes}
RC (horizontal or vertical, depending on plotting convention) measures how close
the mitigated model aligns with the clean reference on contaminated evaluation,
while BUD measures how much uncontaminated performance is lost due to the
mitigation. Points closer to the origin are preferred. A method is Pareto
superior if it achieves lower RC for the same (or lower) BUD.

\paragraph{Role of $\zeta$}
Each point on the DeconIEP curve corresponds to a specific perturbation budget
$\zeta$. Smaller $\zeta$ yields conservative perturbations with low BUD but
limited reduction in RC; larger $\zeta$ increases decontamination strength
(lower RC) at the risk of higher BUD. This
monotone, budget-controlled behavior enables users to select operating points
that match application constraints (e.g., strict preservation of clean accuracy
versus aggressive removal of contamination inflation).

\paragraph{Comparison to baselines}
By plotting baselines alongside the DeconIEP curve, the figure highlights
whether DeconIEP can dominate existing approaches: if baseline points lie above
and/or to the right of the curve, then for the same benign degradation DeconIEP
achieves stronger decontamination, or for the same decontamination level it
incurs less benign harm.

\subsection{Learned perturbations vs.\ random noise}
\label{app:randomnoise}
\begin{figure}[t]
  \centering
  \begin{subfigure}[t]{0.49\linewidth}
    \centering
    \includegraphics[width=\linewidth]{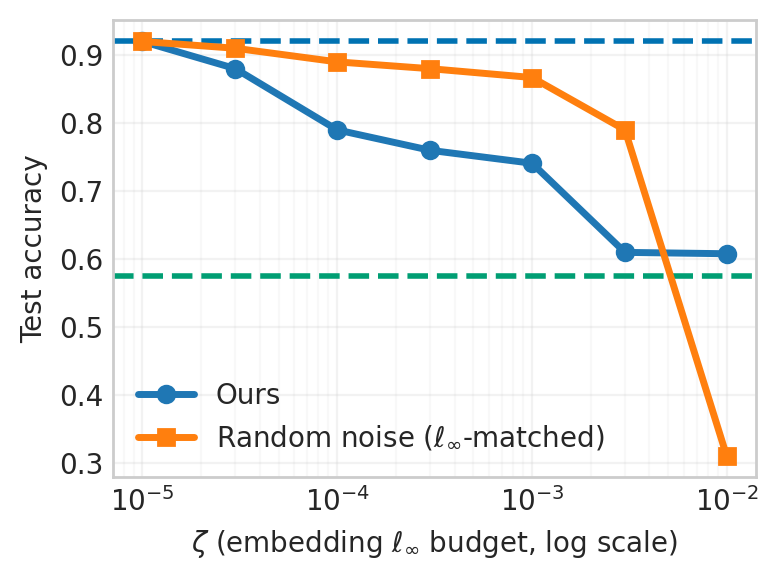}
    \caption{Test accuracy vs.\ $\zeta$ .}
    \label{fig:zeta-acc}
  \end{subfigure}\hfill
  \begin{subfigure}[t]{0.49\linewidth}
    \centering
    \includegraphics[width=\linewidth]{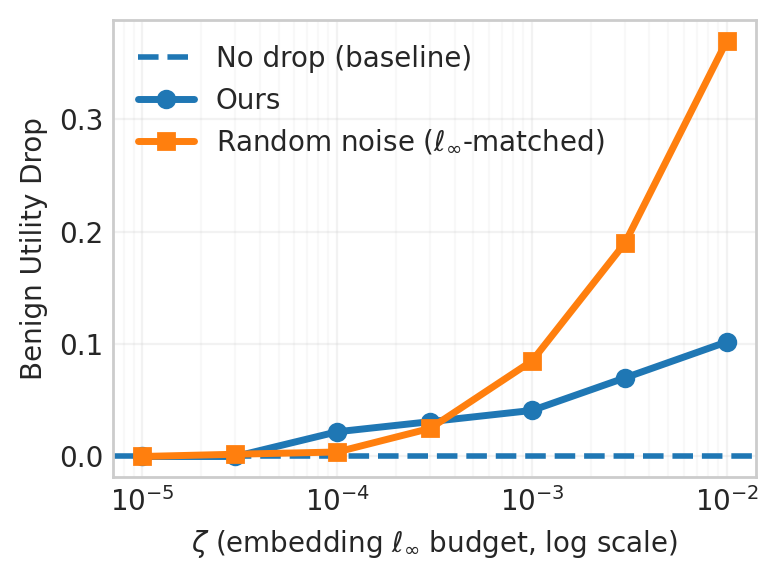}
    \caption{Benign Utility Drop vs.\ $\zeta$ .}
    \label{fig:zeta-cleandrop}
  \end{subfigure}
  \caption{ Learned perturbations vs. random noise across budgets $\zeta$. (a)  The blue and green dashed lines denote the unmitigated contaminated model and the clean-model baseline. (b)  The blue dashed line denotes BUD$=0$.}
  \label{fig:zeta-sweep}
\end{figure}
We test whether the gains come from structured, instance-adaptive
perturbations rather than injecting noise with a comparable magnitude.
We compare {DeconIEP} with an ablation that replaces the learned
perturbation by \textit{random noise} matched in $\ell_\infty$ norm and bounded
by the same $\tanh$ operation.
Figure~\ref{fig:zeta-sweep} shows that random noise is substantially less
reliable: while it can be competitive at very small budgets, its behavior
becomes increasingly unstable as $\zeta$ grows, leading to inconsistent
decontamination and larger BUD. In contrast, DeconIEP improves
decontamination in a smooth and stable manner while incurring a smaller utility
cost, indicating that effective decontamination requires targeted,
instance-adaptive perturbations rather than unstructured noise.
\subsection{Ablation on auxiliary set size}
\label{app:auxsize}
\begin{figure}[t]
  \centering
  \includegraphics[width=0.95\linewidth]{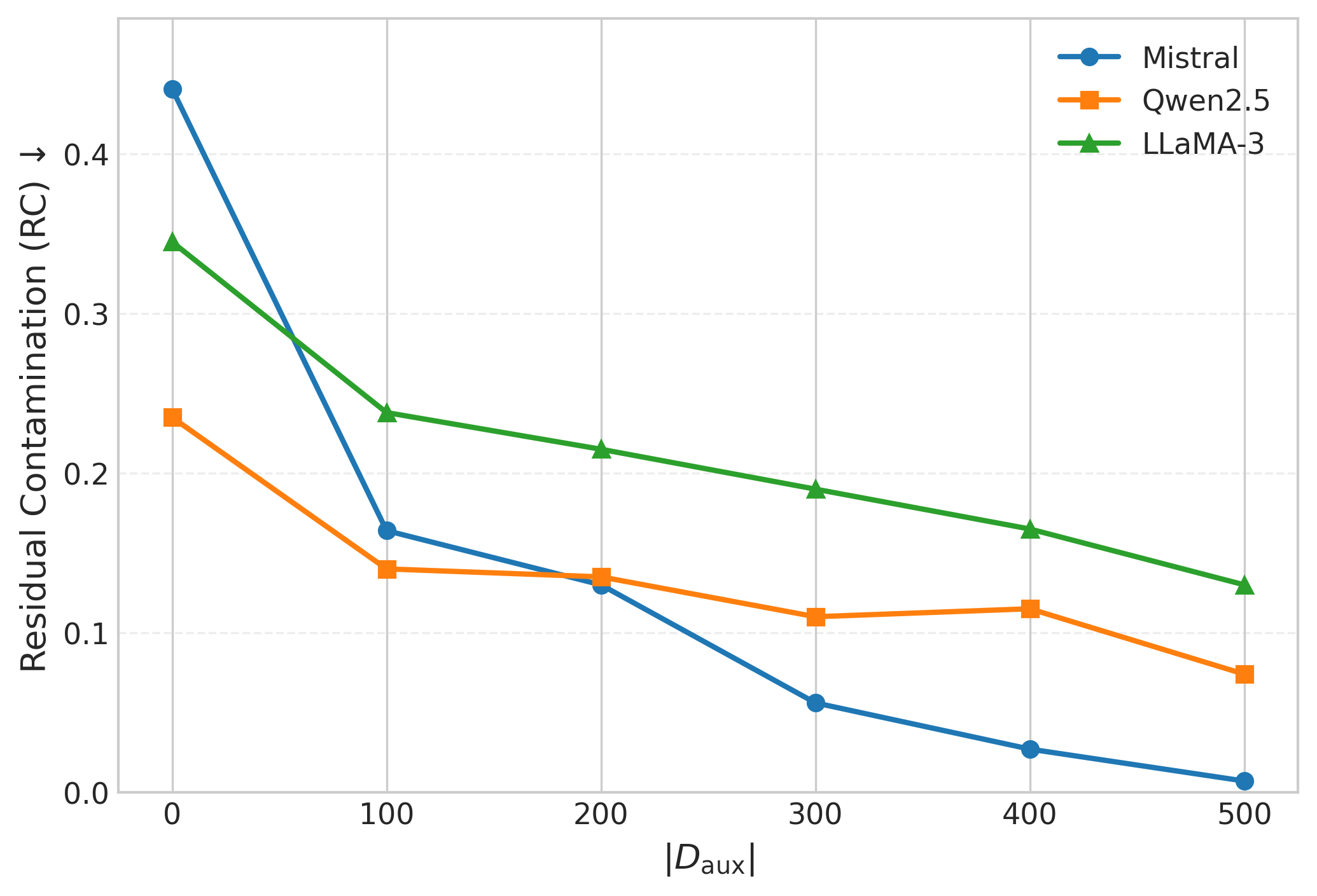}
  \caption{Ablation on the auxiliary set size $|D_{\mathrm{aux}}|$ used to train the perturbation generator. We report Residual Contamination (RC; lower is better) as $|D_{\mathrm{aux}}|$ increases for three backbones (Mistral, Qwen2.5, and LLaMA-3). Increasing $|D_{\mathrm{aux}}|$ consistently reduces RC, indicating that additional auxiliary samples provide a stronger and more stable reference-guided training signal for decontamination.}
  \label{fig:ablation_daux_size}
\end{figure}

We study how the size of the auxiliary dataset $|D_{\mathrm{aux}}|$ affects decontamination performance.
$|D_{\mathrm{aux}}|$ controls how many benchmark samples are used to train the perturbation generator under the reference-guided objective, and thus determines the strength and diversity of the supervision that steers the contaminated model away from memorization-driven behavior.
As shown in Figure~\ref{fig:ablation_daux_size}, RC decreases monotonically (or near-monotonically) as $|D_{\mathrm{aux}}|$ increases across all three backbones, demonstrating that DeconIEP benefits from additional auxiliary samples and that the learned perturbations are not driven by a small set of idiosyncratic examples.
The largest relative improvement typically occurs when moving from $|D_{\mathrm{aux}}|{=}0$ to a small non-zero auxiliary set (e.g., 100), suggesting that even limited auxiliary data can provide sufficient signal to anchor the generator's behavior.
Beyond this regime, further increasing $|D_{\mathrm{aux}}|$ yields diminishing returns, consistent with the intuition that the generator begins to saturate once it has seen enough representative prompts to learn stable, instance-adaptive perturbation patterns.
The trends differ by model family: Mistral shows the steepest early reduction in RC, while Qwen2.5 and LLaMA-3 improve more gradually, suggesting that these backbones may require more diverse auxiliary supervision to consistently suppress contamination-specific shortcuts.
Overall, this ablation supports the practicality of our approach: strong decontamination can be achieved with a relatively small $|D_{\mathrm{aux}}|$, while larger auxiliary sets further improve robustness without changing the discrete prompts or model parameters.

\section{AI Assistant Usage }
We used AI assistants for language polishing and minor code scaffolding; all experimental design, implementation, and results were verified by the authors.
\end{document}